\documentclass[letterpaper, 10 pt, conference]{ieeeconf} 
\IEEEoverridecommandlockouts  
\usepackage{pstricks}
\usepackage{ifplatform}
\usepackage{xkeyval}
\usepackage{rotating}

\usepackage{graphics} 
\usepackage[off]{auto-pst-pdf}
\usepackage{times} 
\usepackage{amsmath} 
\usepackage{psfrag}
\usepackage{cancel}
\usepackage{float}

\usepackage[normalem]{ulem}
\usepackage{cite}
\usepackage{url}
\usepackage{color}
\definecolor{AliceBlue}{rgb}{0.94,0.97,1.00}
\definecolor{AntiqueWhite1}{rgb}{1.00,0.94,0.86}
\definecolor{AntiqueWhite2}{rgb}{0.93,0.87,0.80}
\definecolor{AntiqueWhite3}{rgb}{0.80,0.75,0.69}
\definecolor{AntiqueWhite4}{rgb}{0.55,0.51,0.47}
\definecolor{AntiqueWhite}{rgb}{0.98,0.92,0.84}
\definecolor{BlanchedAlmond}{rgb}{1.00,0.92,0.80}
\definecolor{BlueViolet}{rgb}{0.54,0.17,0.89}
\definecolor{CadetBlue1}{rgb}{0.60,0.96,1.00}
\definecolor{CadetBlue2}{rgb}{0.56,0.90,0.93}
\definecolor{CadetBlue3}{rgb}{0.48,0.77,0.80}
\definecolor{CadetBlue4}{rgb}{0.33,0.53,0.55}
\definecolor{CadetBlue}{rgb}{0.37,0.62,0.63}
\definecolor{CornflowerBlue}{rgb}{0.39,0.58,0.93}
\definecolor{DarkBlue}{rgb}{0.00,0.00,0.55}
\definecolor{DarkCyan}{rgb}{0.00,0.55,0.55}
\definecolor{DarkGoldenrod1}{rgb}{1.00,0.73,0.06}
\definecolor{DarkGoldenrod2}{rgb}{0.93,0.68,0.05}
\definecolor{DarkGoldenrod3}{rgb}{0.80,0.58,0.05}
\definecolor{DarkGoldenrod4}{rgb}{0.55,0.40,0.03}
\definecolor{DarkGoldenrod}{rgb}{0.72,0.53,0.04}
\definecolor{DarkGray}{rgb}{0.66,0.66,0.66}
\definecolor{DarkGreen}{rgb}{0.00,0.39,0.00}
\definecolor{DarkGrey}{rgb}{0.66,0.66,0.66}
\definecolor{DarkKhaki}{rgb}{0.74,0.72,0.42}
\definecolor{DarkMagenta}{rgb}{0.55,0.00,0.55}
\definecolor{DarkOliveGreen1}{rgb}{0.79,1.00,0.44}
\definecolor{DarkOliveGreen2}{rgb}{0.74,0.93,0.41}
\definecolor{DarkOliveGreen3}{rgb}{0.64,0.80,0.35}
\definecolor{DarkOliveGreen4}{rgb}{0.43,0.55,0.24}
\definecolor{DarkOliveGreen}{rgb}{0.33,0.42,0.18}
\definecolor{DarkOrange1}{rgb}{1.00,0.50,0.00}
\definecolor{DarkOrange2}{rgb}{0.93,0.46,0.00}
\definecolor{DarkOrange3}{rgb}{0.80,0.40,0.00}
\definecolor{DarkOrange4}{rgb}{0.55,0.27,0.00}
\definecolor{DarkOrange}{rgb}{1.00,0.55,0.00}
\definecolor{DarkOrchid1}{rgb}{0.75,0.24,1.00}
\definecolor{DarkOrchid2}{rgb}{0.70,0.23,0.93}
\definecolor{DarkOrchid3}{rgb}{0.60,0.20,0.80}
\definecolor{DarkOrchid4}{rgb}{0.41,0.13,0.55}
\definecolor{DarkOrchid}{rgb}{0.60,0.20,0.80}
\definecolor{DarkRed}{rgb}{0.55,0.00,0.00}
\definecolor{DarkSalmon}{rgb}{0.91,0.59,0.48}
\definecolor{DarkSeaGreen1}{rgb}{0.76,1.00,0.76}
\definecolor{DarkSeaGreen2}{rgb}{0.71,0.93,0.71}
\definecolor{DarkSeaGreen3}{rgb}{0.61,0.80,0.61}
\definecolor{DarkSeaGreen4}{rgb}{0.41,0.55,0.41}
\definecolor{DarkSeaGreen}{rgb}{0.56,0.74,0.56}
\definecolor{DarkSlateBlue}{rgb}{0.28,0.24,0.55}
\definecolor{DarkSlateGray1}{rgb}{0.59,1.00,1.00}
\definecolor{DarkSlateGray2}{rgb}{0.55,0.93,0.93}
\definecolor{DarkSlateGray3}{rgb}{0.47,0.80,0.80}
\definecolor{DarkSlateGray4}{rgb}{0.32,0.55,0.55}
\definecolor{DarkSlateGray}{rgb}{0.18,0.31,0.31}
\definecolor{DarkSlateGrey}{rgb}{0.18,0.31,0.31}
\definecolor{DarkTurquoise}{rgb}{0.00,0.81,0.82}
\definecolor{DarkViolet}{rgb}{0.58,0.00,0.83}
\definecolor{DeepPink1}{rgb}{1.00,0.08,0.58}
\definecolor{DeepPink2}{rgb}{0.93,0.07,0.54}
\definecolor{DeepPink3}{rgb}{0.80,0.06,0.46}
\definecolor{DeepPink4}{rgb}{0.55,0.04,0.31}
\definecolor{DeepPink}{rgb}{1.00,0.08,0.58}
\definecolor{DeepSkyBlue1}{rgb}{0.00,0.75,1.00}
\definecolor{DeepSkyBlue2}{rgb}{0.00,0.70,0.93}
\definecolor{DeepSkyBlue3}{rgb}{0.00,0.60,0.80}
\definecolor{DeepSkyBlue4}{rgb}{0.00,0.41,0.55}
\definecolor{DeepSkyBlue}{rgb}{0.00,0.75,1.00}
\definecolor{DimGray}{rgb}{0.41,0.41,0.41}
\definecolor{DimGrey}{rgb}{0.41,0.41,0.41}
\definecolor{DodgerBlue1}{rgb}{0.12,0.56,1.00}
\definecolor{DodgerBlue2}{rgb}{0.11,0.53,0.93}
\definecolor{DodgerBlue3}{rgb}{0.09,0.45,0.80}
\definecolor{DodgerBlue4}{rgb}{0.06,0.31,0.55}
\definecolor{DodgerBlue}{rgb}{0.12,0.56,1.00}
\definecolor{FloralWhite}{rgb}{1.00,0.98,0.94}
\definecolor{ForestGreen}{rgb}{0.13,0.55,0.13}
\definecolor{GhostWhite}{rgb}{0.97,0.97,1.00}
\definecolor{GreenYellow}{rgb}{0.68,1.00,0.18}
\definecolor{HotPink1}{rgb}{1.00,0.43,0.71}
\definecolor{HotPink2}{rgb}{0.93,0.42,0.65}
\definecolor{HotPink3}{rgb}{0.80,0.38,0.56}
\definecolor{HotPink4}{rgb}{0.55,0.23,0.38}
\definecolor{HotPink}{rgb}{1.00,0.41,0.71}
\definecolor{IndianRed1}{rgb}{1.00,0.42,0.42}
\definecolor{IndianRed2}{rgb}{0.93,0.39,0.39}
\definecolor{IndianRed3}{rgb}{0.80,0.33,0.33}
\definecolor{IndianRed4}{rgb}{0.55,0.23,0.23}
\definecolor{IndianRed}{rgb}{0.80,0.36,0.36}
\definecolor{LavenderBlush1}{rgb}{1.00,0.94,0.96}
\definecolor{LavenderBlush2}{rgb}{0.93,0.88,0.90}
\definecolor{LavenderBlush3}{rgb}{0.80,0.76,0.77}
\definecolor{LavenderBlush4}{rgb}{0.55,0.51,0.53}
\definecolor{LavenderBlush}{rgb}{1.00,0.94,0.96}
\definecolor{LawnGreen}{rgb}{0.49,0.99,0.00}
\definecolor{LemonChiffon1}{rgb}{1.00,0.98,0.80}
\definecolor{LemonChiffon2}{rgb}{0.93,0.91,0.75}
\definecolor{LemonChiffon3}{rgb}{0.80,0.79,0.65}
\definecolor{LemonChiffon4}{rgb}{0.55,0.54,0.44}
\definecolor{LemonChiffon}{rgb}{1.00,0.98,0.80}
\definecolor{LightBlue1}{rgb}{0.75,0.94,1.00}
\definecolor{LightBlue2}{rgb}{0.70,0.87,0.93}
\definecolor{LightBlue3}{rgb}{0.60,0.75,0.80}
\definecolor{LightBlue4}{rgb}{0.41,0.51,0.55}
\definecolor{LightBlue}{rgb}{0.68,0.85,0.90}
\definecolor{LightCoral}{rgb}{0.94,0.50,0.50}
\definecolor{LightCyan1}{rgb}{0.88,1.00,1.00}
\definecolor{LightCyan2}{rgb}{0.82,0.93,0.93}
\definecolor{LightCyan3}{rgb}{0.71,0.80,0.80}
\definecolor{LightCyan4}{rgb}{0.48,0.55,0.55}
\definecolor{LightCyan}{rgb}{0.88,1.00,1.00}
\definecolor{LightGoldenrod1}{rgb}{1.00,0.93,0.55}
\definecolor{LightGoldenrod2}{rgb}{0.93,0.86,0.51}
\definecolor{LightGoldenrod3}{rgb}{0.80,0.75,0.44}
\definecolor{LightGoldenrod4}{rgb}{0.55,0.51,0.30}
\definecolor{LightGoldenrodYellow}{rgb}{0.98,0.98,0.82}
\definecolor{LightGoldenrod}{rgb}{0.93,0.87,0.51}
\definecolor{LightGray}{rgb}{0.83,0.83,0.83}
\definecolor{LightGreen}{rgb}{0.56,0.93,0.56}
\definecolor{LightGrey}{rgb}{0.83,0.83,0.83}
\definecolor{LightPink1}{rgb}{1.00,0.68,0.73}
\definecolor{LightPink2}{rgb}{0.93,0.64,0.68}
\definecolor{LightPink3}{rgb}{0.80,0.55,0.58}
\definecolor{LightPink4}{rgb}{0.55,0.37,0.40}
\definecolor{LightPink}{rgb}{1.00,0.71,0.76}
\definecolor{LightSalmon1}{rgb}{1.00,0.63,0.48}
\definecolor{LightSalmon2}{rgb}{0.93,0.58,0.45}
\definecolor{LightSalmon3}{rgb}{0.80,0.51,0.38}
\definecolor{LightSalmon4}{rgb}{0.55,0.34,0.26}
\definecolor{LightSalmon}{rgb}{1.00,0.63,0.48}
\definecolor{LightSeaGreen}{rgb}{0.13,0.70,0.67}
\definecolor{LightSkyBlue1}{rgb}{0.69,0.89,1.00}
\definecolor{LightSkyBlue2}{rgb}{0.64,0.83,0.93}
\definecolor{LightSkyBlue3}{rgb}{0.55,0.71,0.80}
\definecolor{LightSkyBlue4}{rgb}{0.38,0.48,0.55}
\definecolor{LightSkyBlue}{rgb}{0.53,0.81,0.98}
\definecolor{LightSlateBlue}{rgb}{0.52,0.44,1.00}
\definecolor{LightSlateGray}{rgb}{0.47,0.53,0.60}
\definecolor{LightSlateGrey}{rgb}{0.47,0.53,0.60}
\definecolor{LightSteelBlue1}{rgb}{0.79,0.88,1.00}
\definecolor{LightSteelBlue2}{rgb}{0.74,0.82,0.93}
\definecolor{LightSteelBlue3}{rgb}{0.64,0.71,0.80}
\definecolor{LightSteelBlue4}{rgb}{0.43,0.48,0.55}
\definecolor{LightSteelBlue}{rgb}{0.69,0.77,0.87}
\definecolor{LightYellow1}{rgb}{1.00,1.00,0.88}
\definecolor{LightYellow2}{rgb}{0.93,0.93,0.82}
\definecolor{LightYellow3}{rgb}{0.80,0.80,0.71}
\definecolor{LightYellow4}{rgb}{0.55,0.55,0.48}
\definecolor{LightYellow}{rgb}{1.00,1.00,0.88}
\definecolor{LimeGreen}{rgb}{0.20,0.80,0.20}
\definecolor{MediumAquamarine}{rgb}{0.40,0.80,0.67}
\definecolor{MediumBlue}{rgb}{0.00,0.00,0.80}
\definecolor{MediumOrchid1}{rgb}{0.88,0.40,1.00}
\definecolor{MediumOrchid2}{rgb}{0.82,0.37,0.93}
\definecolor{MediumOrchid3}{rgb}{0.71,0.32,0.80}
\definecolor{MediumOrchid4}{rgb}{0.48,0.22,0.55}
\definecolor{MediumOrchid}{rgb}{0.73,0.33,0.83}
\definecolor{MediumPurple1}{rgb}{0.67,0.51,1.00}
\definecolor{MediumPurple2}{rgb}{0.62,0.47,0.93}
\definecolor{MediumPurple3}{rgb}{0.54,0.41,0.80}
\definecolor{MediumPurple4}{rgb}{0.36,0.28,0.55}
\definecolor{MediumPurple}{rgb}{0.58,0.44,0.86}
\definecolor{MediumSeaGreen}{rgb}{0.24,0.70,0.44}
\definecolor{MediumSlateBlue}{rgb}{0.48,0.41,0.93}
\definecolor{MediumSpringGreen}{rgb}{0.00,0.98,0.60}
\definecolor{MediumTurquoise}{rgb}{0.28,0.82,0.80}
\definecolor{MediumVioletRed}{rgb}{0.78,0.08,0.52}
\definecolor{MidnightBlue}{rgb}{0.10,0.10,0.44}
\definecolor{MintCream}{rgb}{0.96,1.00,0.98}
\definecolor{MistyRose1}{rgb}{1.00,0.89,0.88}
\definecolor{MistyRose2}{rgb}{0.93,0.84,0.82}
\definecolor{MistyRose3}{rgb}{0.80,0.72,0.71}
\definecolor{MistyRose4}{rgb}{0.55,0.49,0.48}
\definecolor{MistyRose}{rgb}{1.00,0.89,0.88}
\definecolor{NavajoWhite1}{rgb}{1.00,0.87,0.68}
\definecolor{NavajoWhite2}{rgb}{0.93,0.81,0.63}
\definecolor{NavajoWhite3}{rgb}{0.80,0.70,0.55}
\definecolor{NavajoWhite4}{rgb}{0.55,0.47,0.37}
\definecolor{NavajoWhite}{rgb}{1.00,0.87,0.68}
\definecolor{NavyBlue}{rgb}{0.00,0.00,0.50}
\definecolor{OldLace}{rgb}{0.99,0.96,0.90}
\definecolor{OliveDrab1}{rgb}{0.75,1.00,0.24}
\definecolor{OliveDrab2}{rgb}{0.70,0.93,0.23}
\definecolor{OliveDrab3}{rgb}{0.60,0.80,0.20}
\definecolor{OliveDrab4}{rgb}{0.41,0.55,0.13}
\definecolor{OliveDrab}{rgb}{0.42,0.56,0.14}
\definecolor{OrangeRed1}{rgb}{1.00,0.27,0.00}
\definecolor{OrangeRed2}{rgb}{0.93,0.25,0.00}
\definecolor{OrangeRed3}{rgb}{0.80,0.22,0.00}
\definecolor{OrangeRed4}{rgb}{0.55,0.15,0.00}
\definecolor{OrangeRed}{rgb}{1.00,0.27,0.00}
\definecolor{PaleGoldenrod}{rgb}{0.93,0.91,0.67}
\definecolor{PaleGreen1}{rgb}{0.60,1.00,0.60}
\definecolor{PaleGreen2}{rgb}{0.56,0.93,0.56}
\definecolor{PaleGreen3}{rgb}{0.49,0.80,0.49}
\definecolor{PaleGreen4}{rgb}{0.33,0.55,0.33}
\definecolor{PaleGreen}{rgb}{0.60,0.98,0.60}
\definecolor{PaleTurquoise1}{rgb}{0.73,1.00,1.00}
\definecolor{PaleTurquoise2}{rgb}{0.68,0.93,0.93}
\definecolor{PaleTurquoise3}{rgb}{0.59,0.80,0.80}
\definecolor{PaleTurquoise4}{rgb}{0.40,0.55,0.55}
\definecolor{PaleTurquoise}{rgb}{0.69,0.93,0.93}
\definecolor{PaleVioletRed1}{rgb}{1.00,0.51,0.67}
\definecolor{PaleVioletRed2}{rgb}{0.93,0.47,0.62}
\definecolor{PaleVioletRed3}{rgb}{0.80,0.41,0.54}
\definecolor{PaleVioletRed4}{rgb}{0.55,0.28,0.36}
\definecolor{PaleVioletRed}{rgb}{0.86,0.44,0.58}
\definecolor{PapayaWhip}{rgb}{1.00,0.94,0.84}
\definecolor{PeachPuff1}{rgb}{1.00,0.85,0.73}
\definecolor{PeachPuff2}{rgb}{0.93,0.80,0.68}
\definecolor{PeachPuff3}{rgb}{0.80,0.69,0.58}
\definecolor{PeachPuff4}{rgb}{0.55,0.47,0.40}
\definecolor{PeachPuff}{rgb}{1.00,0.85,0.73}
\definecolor{PowderBlue}{rgb}{0.69,0.88,0.90}
\definecolor{RosyBrown1}{rgb}{1.00,0.76,0.76}
\definecolor{RosyBrown2}{rgb}{0.93,0.71,0.71}
\definecolor{RosyBrown3}{rgb}{0.80,0.61,0.61}
\definecolor{RosyBrown4}{rgb}{0.55,0.41,0.41}
\definecolor{RosyBrown}{rgb}{0.74,0.56,0.56}
\definecolor{RoyalBlue1}{rgb}{0.28,0.46,1.00}
\definecolor{RoyalBlue2}{rgb}{0.26,0.43,0.93}
\definecolor{RoyalBlue3}{rgb}{0.23,0.37,0.80}
\definecolor{RoyalBlue4}{rgb}{0.15,0.25,0.55}
\definecolor{RoyalBlue}{rgb}{0.25,0.41,0.88}
\definecolor{SaddleBrown}{rgb}{0.55,0.27,0.07}
\definecolor{SandyBrown}{rgb}{0.96,0.64,0.38}
\definecolor{SeaGreen1}{rgb}{0.33,1.00,0.62}
\definecolor{SeaGreen2}{rgb}{0.31,0.93,0.58}
\definecolor{SeaGreen3}{rgb}{0.26,0.80,0.50}
\definecolor{SeaGreen4}{rgb}{0.18,0.55,0.34}
\definecolor{SeaGreen}{rgb}{0.18,0.55,0.34}
\definecolor{SkyBlue1}{rgb}{0.53,0.81,1.00}
\definecolor{SkyBlue2}{rgb}{0.49,0.75,0.93}
\definecolor{SkyBlue3}{rgb}{0.42,0.65,0.80}
\definecolor{SkyBlue4}{rgb}{0.29,0.44,0.55}
\definecolor{SkyBlue}{rgb}{0.53,0.81,0.92}
\definecolor{SlateBlue1}{rgb}{0.51,0.44,1.00}
\definecolor{SlateBlue2}{rgb}{0.48,0.40,0.93}
\definecolor{SlateBlue3}{rgb}{0.41,0.35,0.80}
\definecolor{SlateBlue4}{rgb}{0.28,0.24,0.55}
\definecolor{SlateBlue}{rgb}{0.42,0.35,0.80}
\definecolor{SlateGray1}{rgb}{0.78,0.89,1.00}
\definecolor{SlateGray2}{rgb}{0.73,0.83,0.93}
\definecolor{SlateGray3}{rgb}{0.62,0.71,0.80}
\definecolor{SlateGray4}{rgb}{0.42,0.48,0.55}
\definecolor{SlateGray}{rgb}{0.44,0.50,0.56}
\definecolor{SlateGrey}{rgb}{0.44,0.50,0.56}
\definecolor{SpringGreen1}{rgb}{0.00,1.00,0.50}
\definecolor{SpringGreen2}{rgb}{0.00,0.93,0.46}
\definecolor{SpringGreen3}{rgb}{0.00,0.80,0.40}
\definecolor{SpringGreen4}{rgb}{0.00,0.55,0.27}
\definecolor{SpringGreen}{rgb}{0.00,1.00,0.50}
\definecolor{SteelBlue1}{rgb}{0.39,0.72,1.00}
\definecolor{SteelBlue2}{rgb}{0.36,0.67,0.93}
\definecolor{SteelBlue3}{rgb}{0.31,0.58,0.80}
\definecolor{SteelBlue4}{rgb}{0.21,0.39,0.55}
\definecolor{SteelBlue}{rgb}{0.27,0.51,0.71}
\definecolor{VioletRed1}{rgb}{1.00,0.24,0.59}
\definecolor{VioletRed2}{rgb}{0.93,0.23,0.55}
\definecolor{VioletRed3}{rgb}{0.80,0.20,0.47}
\definecolor{VioletRed4}{rgb}{0.55,0.13,0.32}
\definecolor{VioletRed}{rgb}{0.82,0.13,0.56}
\definecolor{WhiteSmoke}{rgb}{0.96,0.96,0.96}
\definecolor{YellowGreen}{rgb}{0.60,0.80,0.20}
\definecolor{aliceblue}{rgb}{0.94,0.97,1.00}
\definecolor{antiquewhite}{rgb}{0.98,0.92,0.84}
\definecolor{aquamarine1}{rgb}{0.50,1.00,0.83}
\definecolor{aquamarine2}{rgb}{0.46,0.93,0.78}
\definecolor{aquamarine3}{rgb}{0.40,0.80,0.67}
\definecolor{aquamarine4}{rgb}{0.27,0.55,0.45}
\definecolor{aquamarine}{rgb}{0.50,1.00,0.83}
\definecolor{azure1}{rgb}{0.94,1.00,1.00}
\definecolor{azure2}{rgb}{0.88,0.93,0.93}
\definecolor{azure3}{rgb}{0.76,0.80,0.80}
\definecolor{azure4}{rgb}{0.51,0.55,0.55}
\definecolor{azure}{rgb}{0.94,1.00,1.00}
\definecolor{beige}{rgb}{0.96,0.96,0.86}
\definecolor{bisque1}{rgb}{1.00,0.89,0.77}
\definecolor{bisque2}{rgb}{0.93,0.84,0.72}
\definecolor{bisque3}{rgb}{0.80,0.72,0.62}
\definecolor{bisque4}{rgb}{0.55,0.49,0.42}
\definecolor{bisque}{rgb}{1.00,0.89,0.77}
\definecolor{black}{rgb}{0.00,0.00,0.00}
\definecolor{blanchedalmond}{rgb}{1.00,0.92,0.80}
\definecolor{blue1}{rgb}{0.00,0.00,1.00}
\definecolor{blue2}{rgb}{0.00,0.00,0.93}
\definecolor{blue3}{rgb}{0.00,0.00,0.80}
\definecolor{blue4}{rgb}{0.00,0.00,0.55}
\definecolor{blueviolet}{rgb}{0.54,0.17,0.89}
\definecolor{blue}{rgb}{0.00,0.00,1.00}
\definecolor{brown1}{rgb}{1.00,0.25,0.25}
\definecolor{brown2}{rgb}{0.93,0.23,0.23}
\definecolor{brown3}{rgb}{0.80,0.20,0.20}
\definecolor{brown4}{rgb}{0.55,0.14,0.14}
\definecolor{brown}{rgb}{0.65,0.16,0.16}
\definecolor{burlywood1}{rgb}{1.00,0.83,0.61}
\definecolor{burlywood2}{rgb}{0.93,0.77,0.57}
\definecolor{burlywood3}{rgb}{0.80,0.67,0.49}
\definecolor{burlywood4}{rgb}{0.55,0.45,0.33}
\definecolor{burlywood}{rgb}{0.87,0.72,0.53}
\definecolor{cadetblue}{rgb}{0.37,0.62,0.63}
\definecolor{chartreuse1}{rgb}{0.50,1.00,0.00}
\definecolor{chartreuse2}{rgb}{0.46,0.93,0.00}
\definecolor{chartreuse3}{rgb}{0.40,0.80,0.00}
\definecolor{chartreuse4}{rgb}{0.27,0.55,0.00}
\definecolor{chartreuse}{rgb}{0.50,1.00,0.00}
\definecolor{chocolate1}{rgb}{1.00,0.50,0.14}
\definecolor{chocolate2}{rgb}{0.93,0.46,0.13}
\definecolor{chocolate3}{rgb}{0.80,0.40,0.11}
\definecolor{chocolate4}{rgb}{0.55,0.27,0.07}
\definecolor{chocolate}{rgb}{0.82,0.41,0.12}
\definecolor{coral1}{rgb}{1.00,0.45,0.34}
\definecolor{coral2}{rgb}{0.93,0.42,0.31}
\definecolor{coral3}{rgb}{0.80,0.36,0.27}
\definecolor{coral4}{rgb}{0.55,0.24,0.18}
\definecolor{coral}{rgb}{1.00,0.50,0.31}
\definecolor{cornflowerblue}{rgb}{0.39,0.58,0.93}
\definecolor{cornsilk1}{rgb}{1.00,0.97,0.86}
\definecolor{cornsilk2}{rgb}{0.93,0.91,0.80}
\definecolor{cornsilk3}{rgb}{0.80,0.78,0.69}
\definecolor{cornsilk4}{rgb}{0.55,0.53,0.47}
\definecolor{cornsilk}{rgb}{1.00,0.97,0.86}
\definecolor{cyan1}{rgb}{0.00,1.00,1.00}
\definecolor{cyan2}{rgb}{0.00,0.93,0.93}
\definecolor{cyan3}{rgb}{0.00,0.80,0.80}
\definecolor{cyan4}{rgb}{0.00,0.55,0.55}
\definecolor{cyan}{rgb}{0.00,1.00,1.00}
\definecolor{darkblue}{rgb}{0.00,0.00,0.55}
\definecolor{darkcyan}{rgb}{0.00,0.55,0.55}
\definecolor{darkgoldenrod}{rgb}{0.72,0.53,0.04}
\definecolor{darkgray}{rgb}{0.66,0.66,0.66}
\definecolor{darkgreen}{rgb}{0.00,0.39,0.00}
\definecolor{darkgrey}{rgb}{0.66,0.66,0.66}
\definecolor{darkkhaki}{rgb}{0.74,0.72,0.42}
\definecolor{darkmagenta}{rgb}{0.55,0.00,0.55}
\definecolor{darkolive}{rgb}{0.33,0.42,0.18}
\definecolor{darkorange}{rgb}{1.00,0.55,0.00}
\definecolor{darkorchid}{rgb}{0.60,0.20,0.80}
\definecolor{darkred}{rgb}{0.55,0.00,0.00}
\definecolor{darksalmon}{rgb}{0.91,0.59,0.48}
\definecolor{darksea}{rgb}{0.56,0.74,0.56}
\definecolor{darkslate}{rgb}{0.18,0.31,0.31}
\definecolor{darkslate}{rgb}{0.18,0.31,0.31}
\definecolor{darkslate}{rgb}{0.28,0.24,0.55}
\definecolor{darkturquoise}{rgb}{0.00,0.81,0.82}
\definecolor{darkviolet}{rgb}{0.58,0.00,0.83}
\definecolor{deeppink}{rgb}{1.00,0.08,0.58}
\definecolor{deepsky}{rgb}{0.00,0.75,1.00}
\definecolor{dimgray}{rgb}{0.41,0.41,0.41}
\definecolor{dimgrey}{rgb}{0.41,0.41,0.41}
\definecolor{dodgerblue}{rgb}{0.12,0.56,1.00}
\definecolor{firebrick1}{rgb}{1.00,0.19,0.19}
\definecolor{firebrick2}{rgb}{0.93,0.17,0.17}
\definecolor{firebrick3}{rgb}{0.80,0.15,0.15}
\definecolor{firebrick4}{rgb}{0.55,0.10,0.10}
\definecolor{firebrick}{rgb}{0.70,0.13,0.13}
\definecolor{floralwhite}{rgb}{1.00,0.98,0.94}
\definecolor{forestgreen}{rgb}{0.13,0.55,0.13}
\definecolor{gainsboro}{rgb}{0.86,0.86,0.86}
\definecolor{ghostwhite}{rgb}{0.97,0.97,1.00}
\definecolor{gold1}{rgb}{1.00,0.84,0.00}
\definecolor{gold2}{rgb}{0.93,0.79,0.00}
\definecolor{gold3}{rgb}{0.80,0.68,0.00}
\definecolor{gold4}{rgb}{0.55,0.46,0.00}
\definecolor{goldenrod1}{rgb}{1.00,0.76,0.15}
\definecolor{goldenrod2}{rgb}{0.93,0.71,0.13}
\definecolor{goldenrod3}{rgb}{0.80,0.61,0.11}
\definecolor{goldenrod4}{rgb}{0.55,0.41,0.08}
\definecolor{goldenrod}{rgb}{0.85,0.65,0.13}
\definecolor{gold}{rgb}{1.00,0.84,0.00}
\definecolor{gray0}{rgb}{0.00,0.00,0.00}
\definecolor{gray100}{rgb}{1.00,1.00,1.00}
\definecolor{gray10}{rgb}{0.10,0.10,0.10}
\definecolor{gray11}{rgb}{0.11,0.11,0.11}
\definecolor{gray12}{rgb}{0.12,0.12,0.12}
\definecolor{gray13}{rgb}{0.13,0.13,0.13}
\definecolor{gray14}{rgb}{0.14,0.14,0.14}
\definecolor{gray15}{rgb}{0.15,0.15,0.15}
\definecolor{gray16}{rgb}{0.16,0.16,0.16}
\definecolor{gray17}{rgb}{0.17,0.17,0.17}
\definecolor{gray18}{rgb}{0.18,0.18,0.18}
\definecolor{gray19}{rgb}{0.19,0.19,0.19}
\definecolor{gray1}{rgb}{0.01,0.01,0.01}
\definecolor{gray20}{rgb}{0.20,0.20,0.20}
\definecolor{gray21}{rgb}{0.21,0.21,0.21}
\definecolor{gray22}{rgb}{0.22,0.22,0.22}
\definecolor{gray23}{rgb}{0.23,0.23,0.23}
\definecolor{gray24}{rgb}{0.24,0.24,0.24}
\definecolor{gray25}{rgb}{0.25,0.25,0.25}
\definecolor{gray26}{rgb}{0.26,0.26,0.26}
\definecolor{gray27}{rgb}{0.27,0.27,0.27}
\definecolor{gray28}{rgb}{0.28,0.28,0.28}
\definecolor{gray29}{rgb}{0.29,0.29,0.29}
\definecolor{gray2}{rgb}{0.02,0.02,0.02}
\definecolor{gray30}{rgb}{0.30,0.30,0.30}
\definecolor{gray31}{rgb}{0.31,0.31,0.31}
\definecolor{gray32}{rgb}{0.32,0.32,0.32}
\definecolor{gray33}{rgb}{0.33,0.33,0.33}
\definecolor{gray34}{rgb}{0.34,0.34,0.34}
\definecolor{gray35}{rgb}{0.35,0.35,0.35}
\definecolor{gray36}{rgb}{0.36,0.36,0.36}
\definecolor{gray37}{rgb}{0.37,0.37,0.37}
\definecolor{gray38}{rgb}{0.38,0.38,0.38}
\definecolor{gray39}{rgb}{0.39,0.39,0.39}
\definecolor{gray3}{rgb}{0.03,0.03,0.03}
\definecolor{gray40}{rgb}{0.40,0.40,0.40}
\definecolor{gray41}{rgb}{0.41,0.41,0.41}
\definecolor{gray42}{rgb}{0.42,0.42,0.42}
\definecolor{gray43}{rgb}{0.43,0.43,0.43}
\definecolor{gray44}{rgb}{0.44,0.44,0.44}
\definecolor{gray45}{rgb}{0.45,0.45,0.45}
\definecolor{gray46}{rgb}{0.46,0.46,0.46}
\definecolor{gray47}{rgb}{0.47,0.47,0.47}
\definecolor{gray48}{rgb}{0.48,0.48,0.48}
\definecolor{gray49}{rgb}{0.49,0.49,0.49}
\definecolor{gray4}{rgb}{0.04,0.04,0.04}
\definecolor{gray50}{rgb}{0.50,0.50,0.50}
\definecolor{gray51}{rgb}{0.51,0.51,0.51}
\definecolor{gray52}{rgb}{0.52,0.52,0.52}
\definecolor{gray53}{rgb}{0.53,0.53,0.53}
\definecolor{gray54}{rgb}{0.54,0.54,0.54}
\definecolor{gray55}{rgb}{0.55,0.55,0.55}
\definecolor{gray56}{rgb}{0.56,0.56,0.56}
\definecolor{gray57}{rgb}{0.57,0.57,0.57}
\definecolor{gray58}{rgb}{0.58,0.58,0.58}
\definecolor{gray59}{rgb}{0.59,0.59,0.59}
\definecolor{gray5}{rgb}{0.05,0.05,0.05}
\definecolor{gray60}{rgb}{0.60,0.60,0.60}
\definecolor{gray61}{rgb}{0.61,0.61,0.61}
\definecolor{gray62}{rgb}{0.62,0.62,0.62}
\definecolor{gray63}{rgb}{0.63,0.63,0.63}
\definecolor{gray64}{rgb}{0.64,0.64,0.64}
\definecolor{gray65}{rgb}{0.65,0.65,0.65}
\definecolor{gray66}{rgb}{0.66,0.66,0.66}
\definecolor{gray67}{rgb}{0.67,0.67,0.67}
\definecolor{gray68}{rgb}{0.68,0.68,0.68}
\definecolor{gray69}{rgb}{0.69,0.69,0.69}
\definecolor{gray6}{rgb}{0.06,0.06,0.06}
\definecolor{gray70}{rgb}{0.70,0.70,0.70}
\definecolor{gray71}{rgb}{0.71,0.71,0.71}
\definecolor{gray72}{rgb}{0.72,0.72,0.72}
\definecolor{gray73}{rgb}{0.73,0.73,0.73}
\definecolor{gray74}{rgb}{0.74,0.74,0.74}
\definecolor{gray75}{rgb}{0.75,0.75,0.75}
\definecolor{gray76}{rgb}{0.76,0.76,0.76}
\definecolor{gray77}{rgb}{0.77,0.77,0.77}
\definecolor{gray78}{rgb}{0.78,0.78,0.78}
\definecolor{gray79}{rgb}{0.79,0.79,0.79}
\definecolor{gray7}{rgb}{0.07,0.07,0.07}
\definecolor{gray80}{rgb}{0.80,0.80,0.80}
\definecolor{gray81}{rgb}{0.81,0.81,0.81}
\definecolor{gray82}{rgb}{0.82,0.82,0.82}
\definecolor{gray83}{rgb}{0.83,0.83,0.83}
\definecolor{gray84}{rgb}{0.84,0.84,0.84}
\definecolor{gray85}{rgb}{0.85,0.85,0.85}
\definecolor{gray86}{rgb}{0.86,0.86,0.86}
\definecolor{gray87}{rgb}{0.87,0.87,0.87}
\definecolor{gray88}{rgb}{0.88,0.88,0.88}
\definecolor{gray89}{rgb}{0.89,0.89,0.89}
\definecolor{gray8}{rgb}{0.08,0.08,0.08}
\definecolor{gray90}{rgb}{0.90,0.90,0.90}
\definecolor{gray91}{rgb}{0.91,0.91,0.91}
\definecolor{gray92}{rgb}{0.92,0.92,0.92}
\definecolor{gray93}{rgb}{0.93,0.93,0.93}
\definecolor{gray94}{rgb}{0.94,0.94,0.94}
\definecolor{gray95}{rgb}{0.95,0.95,0.95}
\definecolor{gray96}{rgb}{0.96,0.96,0.96}
\definecolor{gray97}{rgb}{0.97,0.97,0.97}
\definecolor{gray98}{rgb}{0.98,0.98,0.98}
\definecolor{gray99}{rgb}{0.99,0.99,0.99}
\definecolor{gray9}{rgb}{0.09,0.09,0.09}
\definecolor{gray}{rgb}{0.75,0.75,0.75}
\definecolor{green1}{rgb}{0.00,1.00,0.00}
\definecolor{green2}{rgb}{0.00,0.93,0.00}
\definecolor{green3}{rgb}{0.00,0.80,0.00}
\definecolor{green4}{rgb}{0.00,0.55,0.00}
\definecolor{greenyellow}{rgb}{0.68,1.00,0.18}
\definecolor{green}{rgb}{0.00,1.00,0.00}
\definecolor{grey0}{rgb}{0.00,0.00,0.00}
\definecolor{grey100}{rgb}{1.00,1.00,1.00}
\definecolor{grey10}{rgb}{0.10,0.10,0.10}
\definecolor{grey11}{rgb}{0.11,0.11,0.11}
\definecolor{grey12}{rgb}{0.12,0.12,0.12}
\definecolor{grey13}{rgb}{0.13,0.13,0.13}
\definecolor{grey14}{rgb}{0.14,0.14,0.14}
\definecolor{grey15}{rgb}{0.15,0.15,0.15}
\definecolor{grey16}{rgb}{0.16,0.16,0.16}
\definecolor{grey17}{rgb}{0.17,0.17,0.17}
\definecolor{grey18}{rgb}{0.18,0.18,0.18}
\definecolor{grey19}{rgb}{0.19,0.19,0.19}
\definecolor{grey1}{rgb}{0.01,0.01,0.01}
\definecolor{grey20}{rgb}{0.20,0.20,0.20}
\definecolor{grey21}{rgb}{0.21,0.21,0.21}
\definecolor{grey22}{rgb}{0.22,0.22,0.22}
\definecolor{grey23}{rgb}{0.23,0.23,0.23}
\definecolor{grey24}{rgb}{0.24,0.24,0.24}
\definecolor{grey25}{rgb}{0.25,0.25,0.25}
\definecolor{grey26}{rgb}{0.26,0.26,0.26}
\definecolor{grey27}{rgb}{0.27,0.27,0.27}
\definecolor{grey28}{rgb}{0.28,0.28,0.28}
\definecolor{grey29}{rgb}{0.29,0.29,0.29}
\definecolor{grey2}{rgb}{0.02,0.02,0.02}
\definecolor{grey30}{rgb}{0.30,0.30,0.30}
\definecolor{grey31}{rgb}{0.31,0.31,0.31}
\definecolor{grey32}{rgb}{0.32,0.32,0.32}
\definecolor{grey33}{rgb}{0.33,0.33,0.33}
\definecolor{grey34}{rgb}{0.34,0.34,0.34}
\definecolor{grey35}{rgb}{0.35,0.35,0.35}
\definecolor{grey36}{rgb}{0.36,0.36,0.36}
\definecolor{grey37}{rgb}{0.37,0.37,0.37}
\definecolor{grey38}{rgb}{0.38,0.38,0.38}
\definecolor{grey39}{rgb}{0.39,0.39,0.39}
\definecolor{grey3}{rgb}{0.03,0.03,0.03}
\definecolor{grey40}{rgb}{0.40,0.40,0.40}
\definecolor{grey41}{rgb}{0.41,0.41,0.41}
\definecolor{grey42}{rgb}{0.42,0.42,0.42}
\definecolor{grey43}{rgb}{0.43,0.43,0.43}
\definecolor{grey44}{rgb}{0.44,0.44,0.44}
\definecolor{grey45}{rgb}{0.45,0.45,0.45}
\definecolor{grey46}{rgb}{0.46,0.46,0.46}
\definecolor{grey47}{rgb}{0.47,0.47,0.47}
\definecolor{grey48}{rgb}{0.48,0.48,0.48}
\definecolor{grey49}{rgb}{0.49,0.49,0.49}
\definecolor{grey4}{rgb}{0.04,0.04,0.04}
\definecolor{grey50}{rgb}{0.50,0.50,0.50}
\definecolor{grey51}{rgb}{0.51,0.51,0.51}
\definecolor{grey52}{rgb}{0.52,0.52,0.52}
\definecolor{grey53}{rgb}{0.53,0.53,0.53}
\definecolor{grey54}{rgb}{0.54,0.54,0.54}
\definecolor{grey55}{rgb}{0.55,0.55,0.55}
\definecolor{grey56}{rgb}{0.56,0.56,0.56}
\definecolor{grey57}{rgb}{0.57,0.57,0.57}
\definecolor{grey58}{rgb}{0.58,0.58,0.58}
\definecolor{grey59}{rgb}{0.59,0.59,0.59}
\definecolor{grey5}{rgb}{0.05,0.05,0.05}
\definecolor{grey60}{rgb}{0.60,0.60,0.60}
\definecolor{grey61}{rgb}{0.61,0.61,0.61}
\definecolor{grey62}{rgb}{0.62,0.62,0.62}
\definecolor{grey63}{rgb}{0.63,0.63,0.63}
\definecolor{grey64}{rgb}{0.64,0.64,0.64}
\definecolor{grey65}{rgb}{0.65,0.65,0.65}
\definecolor{grey66}{rgb}{0.66,0.66,0.66}
\definecolor{grey67}{rgb}{0.67,0.67,0.67}
\definecolor{grey68}{rgb}{0.68,0.68,0.68}
\definecolor{grey69}{rgb}{0.69,0.69,0.69}
\definecolor{grey6}{rgb}{0.06,0.06,0.06}
\definecolor{grey70}{rgb}{0.70,0.70,0.70}
\definecolor{grey71}{rgb}{0.71,0.71,0.71}
\definecolor{grey72}{rgb}{0.72,0.72,0.72}
\definecolor{grey73}{rgb}{0.73,0.73,0.73}
\definecolor{grey74}{rgb}{0.74,0.74,0.74}
\definecolor{grey75}{rgb}{0.75,0.75,0.75}
\definecolor{grey76}{rgb}{0.76,0.76,0.76}
\definecolor{grey77}{rgb}{0.77,0.77,0.77}
\definecolor{grey78}{rgb}{0.78,0.78,0.78}
\definecolor{grey79}{rgb}{0.79,0.79,0.79}
\definecolor{grey7}{rgb}{0.07,0.07,0.07}
\definecolor{grey80}{rgb}{0.80,0.80,0.80}
\definecolor{grey81}{rgb}{0.81,0.81,0.81}
\definecolor{grey82}{rgb}{0.82,0.82,0.82}
\definecolor{grey83}{rgb}{0.83,0.83,0.83}
\definecolor{grey84}{rgb}{0.84,0.84,0.84}
\definecolor{grey85}{rgb}{0.85,0.85,0.85}
\definecolor{grey86}{rgb}{0.86,0.86,0.86}
\definecolor{grey87}{rgb}{0.87,0.87,0.87}
\definecolor{grey88}{rgb}{0.88,0.88,0.88}
\definecolor{grey89}{rgb}{0.89,0.89,0.89}
\definecolor{grey8}{rgb}{0.08,0.08,0.08}
\definecolor{grey90}{rgb}{0.90,0.90,0.90}
\definecolor{grey91}{rgb}{0.91,0.91,0.91}
\definecolor{grey92}{rgb}{0.92,0.92,0.92}
\definecolor{grey93}{rgb}{0.93,0.93,0.93}
\definecolor{grey94}{rgb}{0.94,0.94,0.94}
\definecolor{grey95}{rgb}{0.95,0.95,0.95}
\definecolor{grey96}{rgb}{0.96,0.96,0.96}
\definecolor{grey97}{rgb}{0.97,0.97,0.97}
\definecolor{grey98}{rgb}{0.98,0.98,0.98}
\definecolor{grey99}{rgb}{0.99,0.99,0.99}
\definecolor{grey9}{rgb}{0.09,0.09,0.09}
\definecolor{grey}{rgb}{0.75,0.75,0.75}
\definecolor{honeydew1}{rgb}{0.94,1.00,0.94}
\definecolor{honeydew2}{rgb}{0.88,0.93,0.88}
\definecolor{honeydew3}{rgb}{0.76,0.80,0.76}
\definecolor{honeydew4}{rgb}{0.51,0.55,0.51}
\definecolor{honeydew}{rgb}{0.94,1.00,0.94}
\definecolor{hotpink}{rgb}{1.00,0.41,0.71}
\definecolor{indianred}{rgb}{0.80,0.36,0.36}
\definecolor{ivory1}{rgb}{1.00,1.00,0.94}
\definecolor{ivory2}{rgb}{0.93,0.93,0.88}
\definecolor{ivory3}{rgb}{0.80,0.80,0.76}
\definecolor{ivory4}{rgb}{0.55,0.55,0.51}
\definecolor{ivory}{rgb}{1.00,1.00,0.94}
\definecolor{khaki1}{rgb}{1.00,0.96,0.56}
\definecolor{khaki2}{rgb}{0.93,0.90,0.52}
\definecolor{khaki3}{rgb}{0.80,0.78,0.45}
\definecolor{khaki4}{rgb}{0.55,0.53,0.31}
\definecolor{khaki}{rgb}{0.94,0.90,0.55}
\definecolor{lavenderblush}{rgb}{1.00,0.94,0.96}
\definecolor{lavender}{rgb}{0.90,0.90,0.98}
\definecolor{lawngreen}{rgb}{0.49,0.99,0.00}
\definecolor{lemonchiffon}{rgb}{1.00,0.98,0.80}
\definecolor{lightblue}{rgb}{0.68,0.85,0.90}
\definecolor{lightcoral}{rgb}{0.94,0.50,0.50}
\definecolor{lightcyan}{rgb}{0.88,1.00,1.00}
\definecolor{lightgoldenrod}{rgb}{0.93,0.87,0.51}
\definecolor{lightgoldenrod}{rgb}{0.98,0.98,0.82}
\definecolor{lightgray}{rgb}{0.83,0.83,0.83}
\definecolor{lightgreen}{rgb}{0.56,0.93,0.56}
\definecolor{lightgrey}{rgb}{0.83,0.83,0.83}
\definecolor{lightpink}{rgb}{1.00,0.71,0.76}
\definecolor{lightsalmon}{rgb}{1.00,0.63,0.48}
\definecolor{lightsea}{rgb}{0.13,0.70,0.67}
\definecolor{lightsky}{rgb}{0.53,0.81,0.98}
\definecolor{lightslate}{rgb}{0.47,0.53,0.60}
\definecolor{lightslate}{rgb}{0.47,0.53,0.60}
\definecolor{lightslate}{rgb}{0.52,0.44,1.00}
\definecolor{lightsteel}{rgb}{0.69,0.77,0.87}
\definecolor{lightyellow}{rgb}{1.00,1.00,0.88}
\definecolor{limegreen}{rgb}{0.20,0.80,0.20}
\definecolor{linen}{rgb}{0.98,0.94,0.90}
\definecolor{magenta1}{rgb}{1.00,0.00,1.00}
\definecolor{magenta2}{rgb}{0.93,0.00,0.93}
\definecolor{magenta3}{rgb}{0.80,0.00,0.80}
\definecolor{magenta4}{rgb}{0.55,0.00,0.55}
\definecolor{magenta}{rgb}{1.00,0.00,1.00}
\definecolor{maroon1}{rgb}{1.00,0.20,0.70}
\definecolor{maroon2}{rgb}{0.93,0.19,0.65}
\definecolor{maroon3}{rgb}{0.80,0.16,0.56}
\definecolor{maroon4}{rgb}{0.55,0.11,0.38}
\definecolor{maroon}{rgb}{0.69,0.19,0.38}
\definecolor{mediumaquamarine}{rgb}{0.40,0.80,0.67}
\definecolor{mediumblue}{rgb}{0.00,0.00,0.80}
\definecolor{mediumorchid}{rgb}{0.73,0.33,0.83}
\definecolor{mediumpurple}{rgb}{0.58,0.44,0.86}
\definecolor{mediumsea}{rgb}{0.24,0.70,0.44}
\definecolor{mediumslate}{rgb}{0.48,0.41,0.93}
\definecolor{mediumspring}{rgb}{0.00,0.98,0.60}
\definecolor{mediumturquoise}{rgb}{0.28,0.82,0.80}
\definecolor{mediumviolet}{rgb}{0.78,0.08,0.52}
\definecolor{midnightblue}{rgb}{0.10,0.10,0.44}
\definecolor{mintcream}{rgb}{0.96,1.00,0.98}
\definecolor{mistyrose}{rgb}{1.00,0.89,0.88}
\definecolor{moccasin}{rgb}{1.00,0.89,0.71}
\definecolor{navajowhite}{rgb}{1.00,0.87,0.68}
\definecolor{navyblue}{rgb}{0.00,0.00,0.50}
\definecolor{navy}{rgb}{0.00,0.00,0.50}
\definecolor{oldlace}{rgb}{0.99,0.96,0.90}
\definecolor{olivedrab}{rgb}{0.42,0.56,0.14}
\definecolor{orange1}{rgb}{1.00,0.65,0.00}
\definecolor{orange2}{rgb}{0.93,0.60,0.00}
\definecolor{orange3}{rgb}{0.80,0.52,0.00}
\definecolor{orange4}{rgb}{0.55,0.35,0.00}
\definecolor{orangered}{rgb}{1.00,0.27,0.00}
\definecolor{orange}{rgb}{1.00,0.65,0.00}
\definecolor{orchid1}{rgb}{1.00,0.51,0.98}
\definecolor{orchid2}{rgb}{0.93,0.48,0.91}
\definecolor{orchid3}{rgb}{0.80,0.41,0.79}
\definecolor{orchid4}{rgb}{0.55,0.28,0.54}
\definecolor{orchid}{rgb}{0.85,0.44,0.84}
\definecolor{palegoldenrod}{rgb}{0.93,0.91,0.67}
\definecolor{palegreen}{rgb}{0.60,0.98,0.60}
\definecolor{paleturquoise}{rgb}{0.69,0.93,0.93}
\definecolor{paleviolet}{rgb}{0.86,0.44,0.58}
\definecolor{papayawhip}{rgb}{1.00,0.94,0.84}
\definecolor{peachpuff}{rgb}{1.00,0.85,0.73}
\definecolor{peru}{rgb}{0.80,0.52,0.25}
\definecolor{pink1}{rgb}{1.00,0.71,0.77}
\definecolor{pink2}{rgb}{0.93,0.66,0.72}
\definecolor{pink3}{rgb}{0.80,0.57,0.62}
\definecolor{pink4}{rgb}{0.55,0.39,0.42}
\definecolor{pink}{rgb}{1.00,0.75,0.80}
\definecolor{plum1}{rgb}{1.00,0.73,1.00}
\definecolor{plum2}{rgb}{0.93,0.68,0.93}
\definecolor{plum3}{rgb}{0.80,0.59,0.80}
\definecolor{plum4}{rgb}{0.55,0.40,0.55}
\definecolor{plum}{rgb}{0.87,0.63,0.87}
\definecolor{powderblue}{rgb}{0.69,0.88,0.90}
\definecolor{purple1}{rgb}{0.61,0.19,1.00}
\definecolor{purple2}{rgb}{0.57,0.17,0.93}
\definecolor{purple3}{rgb}{0.49,0.15,0.80}
\definecolor{purple4}{rgb}{0.33,0.10,0.55}
\definecolor{purple}{rgb}{0.63,0.13,0.94}
\definecolor{red1}{rgb}{1.00,0.00,0.00}
\definecolor{red2}{rgb}{0.93,0.00,0.00}
\definecolor{red3}{rgb}{0.80,0.00,0.00}
\definecolor{red4}{rgb}{0.55,0.00,0.00}
\definecolor{red}{rgb}{1.00,0.00,0.00}
\definecolor{rosybrown}{rgb}{0.74,0.56,0.56}
\definecolor{royalblue}{rgb}{0.25,0.41,0.88}
\definecolor{saddlebrown}{rgb}{0.55,0.27,0.07}
\definecolor{salmon1}{rgb}{1.00,0.55,0.41}
\definecolor{salmon2}{rgb}{0.93,0.51,0.38}
\definecolor{salmon3}{rgb}{0.80,0.44,0.33}
\definecolor{salmon4}{rgb}{0.55,0.30,0.22}
\definecolor{salmon}{rgb}{0.98,0.50,0.45}
\definecolor{sandybrown}{rgb}{0.96,0.64,0.38}
\definecolor{seagreen}{rgb}{0.18,0.55,0.34}
\definecolor{seashell1}{rgb}{1.00,0.96,0.93}
\definecolor{seashell2}{rgb}{0.93,0.90,0.87}
\definecolor{seashell3}{rgb}{0.80,0.77,0.75}
\definecolor{seashell4}{rgb}{0.55,0.53,0.51}
\definecolor{seashell}{rgb}{1.00,0.96,0.93}
\definecolor{sienna1}{rgb}{1.00,0.51,0.28}
\definecolor{sienna2}{rgb}{0.93,0.47,0.26}
\definecolor{sienna3}{rgb}{0.80,0.41,0.22}
\definecolor{sienna4}{rgb}{0.55,0.28,0.15}
\definecolor{sienna}{rgb}{0.63,0.32,0.18}
\definecolor{skyblue}{rgb}{0.53,0.81,0.92}
\definecolor{slateblue}{rgb}{0.42,0.35,0.80}
\definecolor{slategray}{rgb}{0.44,0.50,0.56}
\definecolor{slategrey}{rgb}{0.44,0.50,0.56}
\definecolor{snow1}{rgb}{1.00,0.98,0.98}
\definecolor{snow2}{rgb}{0.93,0.91,0.91}
\definecolor{snow3}{rgb}{0.80,0.79,0.79}
\definecolor{snow4}{rgb}{0.55,0.54,0.54}
\definecolor{snow}{rgb}{1.00,0.98,0.98}
\definecolor{springgreen}{rgb}{0.00,1.00,0.50}
\definecolor{steelblue}{rgb}{0.27,0.51,0.71}
\definecolor{tan1}{rgb}{1.00,0.65,0.31}
\definecolor{tan2}{rgb}{0.93,0.60,0.29}
\definecolor{tan3}{rgb}{0.80,0.52,0.25}
\definecolor{tan4}{rgb}{0.55,0.35,0.17}
\definecolor{tan}{rgb}{0.82,0.71,0.55}
\definecolor{thistle1}{rgb}{1.00,0.88,1.00}
\definecolor{thistle2}{rgb}{0.93,0.82,0.93}
\definecolor{thistle3}{rgb}{0.80,0.71,0.80}
\definecolor{thistle4}{rgb}{0.55,0.48,0.55}
\definecolor{thistle}{rgb}{0.85,0.75,0.85}
\definecolor{tomato1}{rgb}{1.00,0.39,0.28}
\definecolor{tomato2}{rgb}{0.93,0.36,0.26}
\definecolor{tomato3}{rgb}{0.80,0.31,0.22}
\definecolor{tomato4}{rgb}{0.55,0.21,0.15}
\definecolor{tomato}{rgb}{1.00,0.39,0.28}
\definecolor{turquoise1}{rgb}{0.00,0.96,1.00}
\definecolor{turquoise2}{rgb}{0.00,0.90,0.93}
\definecolor{turquoise3}{rgb}{0.00,0.77,0.80}
\definecolor{turquoise4}{rgb}{0.00,0.53,0.55}
\definecolor{turquoise}{rgb}{0.25,0.88,0.82}
\definecolor{violetred}{rgb}{0.82,0.13,0.56}
\definecolor{violet}{rgb}{0.93,0.51,0.93}
\definecolor{wheat1}{rgb}{1.00,0.91,0.73}
\definecolor{wheat2}{rgb}{0.93,0.85,0.68}
\definecolor{wheat3}{rgb}{0.80,0.73,0.59}
\definecolor{wheat4}{rgb}{0.55,0.49,0.40}
\definecolor{wheat}{rgb}{0.96,0.87,0.70}
\definecolor{whitesmoke}{rgb}{0.96,0.96,0.96}
\definecolor{white}{rgb}{1.00,1.00,1.00}
\definecolor{yellow1}{rgb}{1.00,1.00,0.00}
\definecolor{yellow2}{rgb}{0.93,0.93,0.00}
\definecolor{yellow3}{rgb}{0.80,0.80,0.00}
\definecolor{yellow4}{rgb}{0.55,0.55,0.00}
\definecolor{yellowgreen}{rgb}{0.60,0.80,0.20}
\definecolor{yellow}{rgb}{1.00,1.00,0.00}

\usepackage{comment}
\usepackage{amssymb}
\usepackage{amsfonts}
\usepackage{algorithmic}
\usepackage{algorithm}
\usepackage{pifont}
\usepackage{booktabs}
\usepackage{lipsum}
\usepackage{array}
\newcolumntype{P}[1]{>{\centering\arraybackslash}p{#1}}

\newcommand{\LP}[1]{\textcolor{red}{}}

\newcommand{\longonly}[1]{}

\usepackage{adjustbox}

\usepackage{subfigure}

\newcolumntype{R}[2]{%
    >{\adjustbox{angle=#1,lap=\width-(#2)}\bgroup}%
    l%
    <{\egroup}%
}

\graphicspath{{figures/},{figures/new_diagrams/}}

\usepackage{hyperref}

\allowdisplaybreaks[2]
\renewcommand{\baselinestretch}{0.99}
\usepackage{caption}

\usepackage{mdwlist}

\let\itemize\undefined
\makecompactlist{itemize}{stditemize}

\let\enumerate\undefined
\makecompactlist{enumerate}{stdenumerate}

\usepackage[tracking=false,kerning=true,spacing=true]{microtype}
\begin{document}

\title{\LARGE \bf Duckiefloat: a Collision-Tolerant Resource-Constrained Blimp for Long-Term Autonomy in Subterranean Environments
\\\rule{1pt}{0pt}}

\author{Yi-Wei Huang$^{*}$, Chen-Lung Lu$^{*}$, Kuan-Lin Chen, Po-Sheng Ser, Jui-Te Huang, Yu-Chia Shen, \\ Pin-Wei Chen, Po-Kai Chang, Sheng-Cheng Lee, Hsueh-Cheng Wang
\thanks{*Y.-W. Huang and C.-L. Lu contributed equally to this work. All authors are with National Chiao Tung University, Taiwan. Corresponding author email:
        {\tt\small hchengwang@g2.nctu.edu.tw}}%
}


\maketitle
\thispagestyle{empty}
\pagestyle{empty}

\begin{abstract}

There are several challenges for search and rescue robots: mobility, perception, autonomy, and communication. Inspired by the DARPA Subterranean (SubT) Challenge, we propose an autonomous blimp robot, which has the advantages of low power consumption and collision-tolerance compared to other aerial vehicles like drones. This is important for search and rescue tasks that usually last for one or more hours. However, the underground constrained passages limit the size of blimp envelope and its payload, making the proposed system resource-constrained. Therefore, a careful design consideration is needed to build a blimp system with on-board artifact search and SLAM. In order to reach long-term operation, a failure-aware algorithm with minimal communication to human supervisor to have situational awareness and send control signals to the blimp when needed.  

We carry out experiments in a controlled environment with the blimp's pose tracked when performing trajectory following in remote control and autonomous scenarios. Artifact search and communication performances are evaluated in real environments. Finally, we show that the ``situational awareness'' human intervention under limited communication allows the blimp to escape from getting stuck (such as in a constrained passage) and prolong the search operation time. Lesson learnt is discussed for future develops of the proposed blimp in different subterranean environments. 

\end{abstract}

%


\section{Introduction}
\label{sec:Introduction}

\subsection{Motivation} \label{Motivation}

\begin{figure}[t]
\includegraphics[width=1.0\columnwidth]{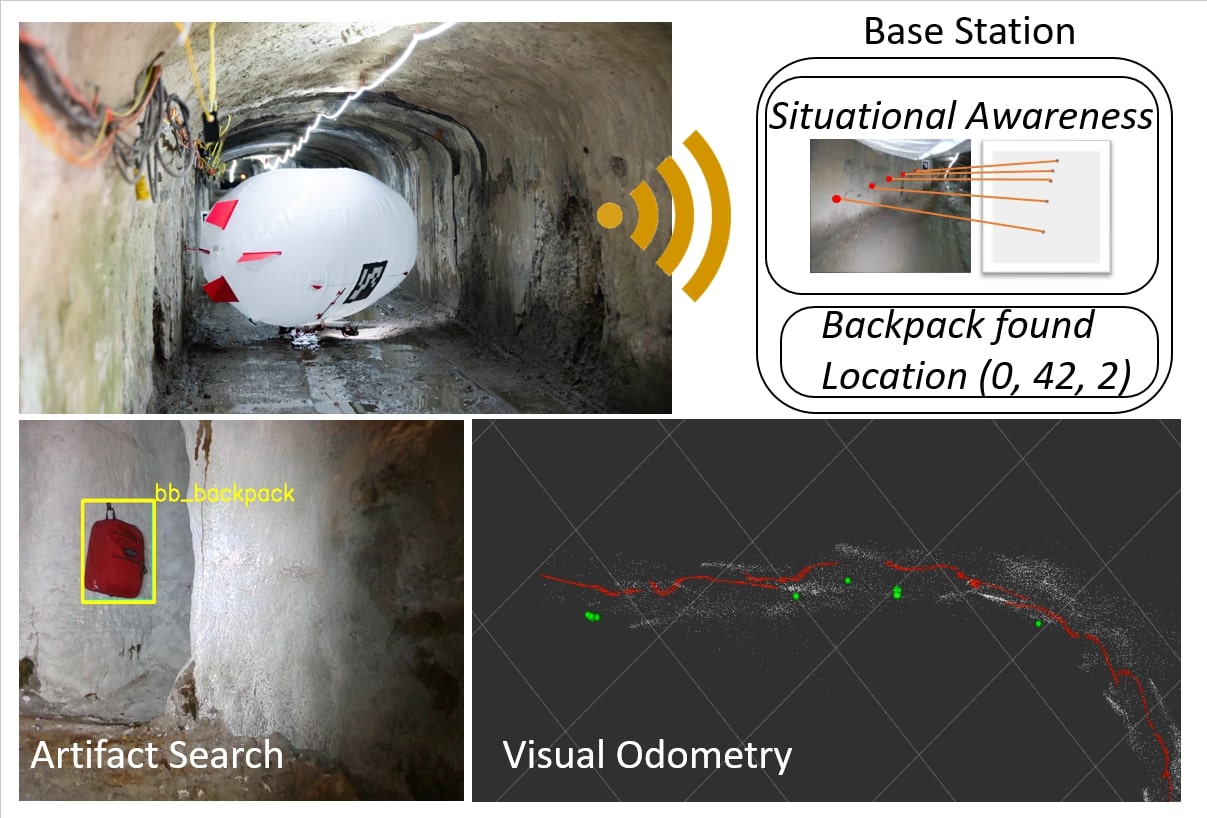}
\centering
\caption{We present an autonomous blimp Duckiefloat for the use of search and rescue (SAR) mission. Compared to quadcopter drone, our blimp is able to achieve longer flight time and is collision-tolerant. The proposed system enables on-board artifact search, visual odometry, and communications to base station for artifact report and situational awareness. Duckiefloat was used in the tunnel circuit of the DARPA Subterranean Challenge.}
\label{fig:teaser}
\end{figure}

Blimps are widely used in the 1900's after its first invented. It was used both in civilian and military transportation needs. But after the appearance of airplanes, most of them are replaced by airplanes due to the safety of using hydrogen and the high cost of helium gas. 

Recent success of unmanned aerial vehicles (UAVs) of quadcopter drones (referred to drones) has make them the most popular UAVs in recent years. It's capable of maneuvering fast and precisely and can carry up to several kilograms in a comparably small body frame. However, the main drawback of drones is that they are power hungry. Drones have only limited flight time (16-40 minutes for a commercially available drone~\cite{sa2017build}), and are prone to collisions and will fail completely if one of the rotor blades are non-functional. Such constraints makes drones limited for search and rescue missions, which take a couple of hours or even days. 

\begin{figure*}[!h]
\includegraphics[width=0.8\textwidth]{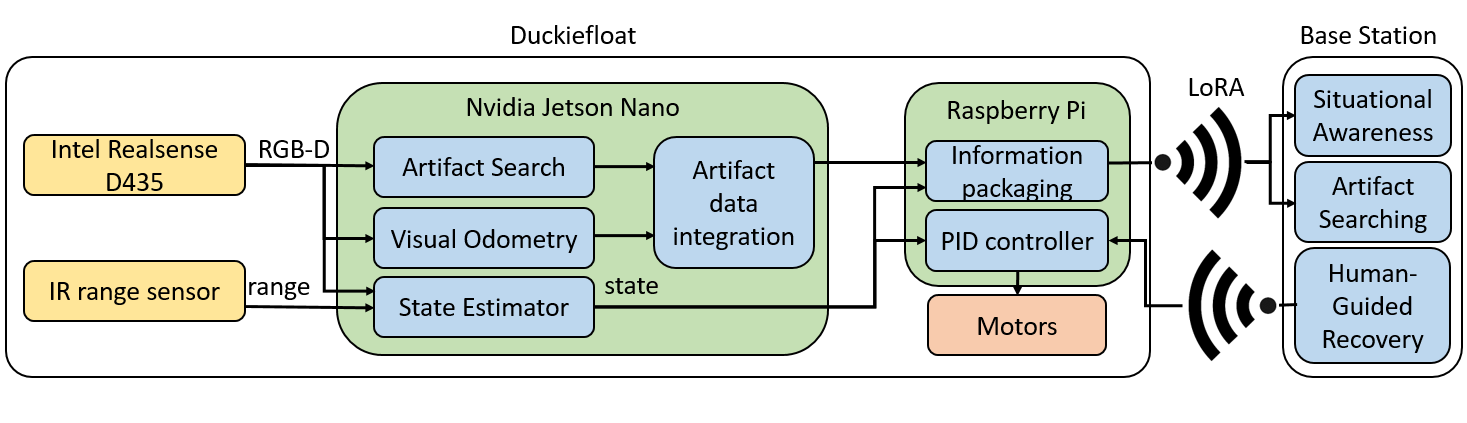}
\centering
\caption{System Overview of Duckiefloat.}
\label{fig:software}
\vspace{-20pt}
\end{figure*}

Blimps have some advantages over drones, and the main upper hand is that it has low-power consumption and is collision-tolerant. 
It is possible that blimps operate for days and even for weeks in a single charge due to its lighter than air (LAT) characteristics. They are also safe to navigate in unstructured environments. Some collisions aren't a big deal due to its non-rigid body. This is important for SAR tasks because it needs to be robust and able to recover from any failure cases. Therefore it is well-suited for the search and rescue contexts, such as in the DARPA Subterranean (SubT) Challenge~\cite{SubT-website}. The main focus of SubT challenge is to encourage robotic research for autonomous systems in subterranean environments. Teams have to finish the task of searching and reporting locations of several types of artifacts including survivors and several items that commonly used by explorers.

Deploying the autonomous blimps as scouting vehicles is a good choice from the operating-time perspective as well as more collision-tolerant than drones in underground environments. Our main goal is to build an autonomous blimp that serves as a scouting robot in a multi-robot system for completing the tasks in the DARPA SubT Challenge, as shown in Fig.~\ref{fig:teaser}. Due to some constrained passages in subterranean environments, blimps have a size limit which is directly related to the weight limit of the payload. Therefore, Duckiefloat is also resource constrained without full-sized computers and heavy sensors. A careful design consideration is needed and will be discussed in Sec.~\ref{system-descriptions}.


This work contributes the follows:

\begin{itemize}

\item \emph{We realized the idea of long-term autonomy via developing a blimp robot ``Duckiefloat'' with the ability of performing SLAM and artifacts search via deep learning approach.}

\item \emph{We implemented a module that provides human supervisors situational awareness and failure recovery mechanism by human if needed.}

\item \emph{We tested Duckiefloat in real tunnel environments, and refined the experience as the future directions of search-purposed blimp research.}

\end{itemize}

\section{Related Work}


\subsection{Search and Rescue (SAR) and Underground Robots}


Autonomous systems will play an essential role in search and rescue applications. Currently, most of the jobs are still done by human beings because in many disaster response cases the sites are unstructured with a lot of debris. They are challenging for robots to perform autonomous tasks. Different types of robots (air, ground, water) have been deployed for search and rescue missions~\cite{delmerico2019current}.
Recent research on search and rescue robots has been also motivated by building robot systems that can actually provide help in natural disasters, such as earthquakes. These robot systems have to tackle challenges including: navigating through unknown and unstructured environments, bad communication, lack of illumination. Some robots~\cite{wang2014development} are designed as ground vehicles with special tracks and suspension systems. This allows them to tackle the mobility problems and adapt to rough terrains. There are some attempts of including autonomy in robotic systems for several years. Gu et al. built a robot system for returning samples from a large outdoor environment and won NASA's Sample Return Robot Challenge in 2014, 2015, 2016 ~\cite{gu2018robot}. Nirouni et al.~\cite{niroui2019deep} used a deep reinforcement learning method for robot search and rescue applications in unknown cluttered environment. However, most of other works are tele-operated, and successful real-world demonstrations of autonomous robots are still rare.



Similarly, the robotic systems operating in subterranean environments have several challenges. There are pipe inspection robots~\cite{montero2015past} designed to detect tunnel defects like erosion and cracks on artificial structures. However due to challenges on autonomy, most robots are tele-operated and still needed operators/workers on-site, which makes operators exposed in danger. TunConstruct~\cite{chmelina2007development} and ROBINSPECT~\cite{loupos2014robotic} are two of the few pipe inspection robotic systems that is close to full autonomy. ROBINSPECT uses laser scanners, ultrasonic sensors, cameras and a robotic arm to operate navigation and inspection tasks. There are also a few research about robot systems in underground mines. ~\cite{roberts2000autonomous} proposed a reactive autonomous navigation algorithm to perform lane following in a mine tunnel without a map.
~\cite{shaffer1992robotic} designed an autonomous mining robot using a 2D laser scanner and triangulation to estimate its pose.

Another big challenge for those search and rescue robot systems in underground environments is localization. The GPS–denied and visually degraded environment caused several common GPS/vision based localization algorithms unusable. State-of-the-art SLAM (Simultaneous Localisation and Mapping) algorithm ORB-SLAM~\cite{mur2017orb} has been shown to perform displeased in underground mine site environments~\cite{jacobson2018semi}. Zeng at el. ~\cite{zeng2019lookup} proposed an visual-based localization algorithm using optical flow and a pixel correspondence match to reject outliers. Huber et al.~\cite{huber2003automatic} used 3D LiDAR mounted on a cart to perform 3D underground mine mapping. Using a graph optimization with a global consistency measure to detect and avoid incorrect, but locally consistent matches. Xiong et al.~\cite{xiong2009integrated} designed an integrated localization system for robots in underground environments using only IMUs. However, the problems of autonomy, perception, communication, and mobility are still challenging, which motivates the DARPA SubT Challenge\cite{SubT-website} we are participating.

\subsection{Autonomous Blimp Systems}

Our proposed system to the DARPA SubT Challenge is an autonomous blimp, which is low-power consumption due to its non-rigid body and the energy-free floating (lighter than air, LAT) characteristic, making it perfectly suited for search and rescue tasks. Several researches about blimp control using model-based methods~\cite{gonzalez2009developing}, ~\cite{fedorenko2016indoor} and using model-free reinforcement learning ~\cite{rottmann2007towards}, ~\cite{ko2007gaussian}. Vision-based SLAM is also implemented on blimps~\cite{hygounenc2004autonomous}. Applications like surveillance system have been built by using visual tracking technologies~\cite{fukao2003autonomous}.

\begin{figure}[!hh]
\includegraphics[width=0.8\columnwidth]{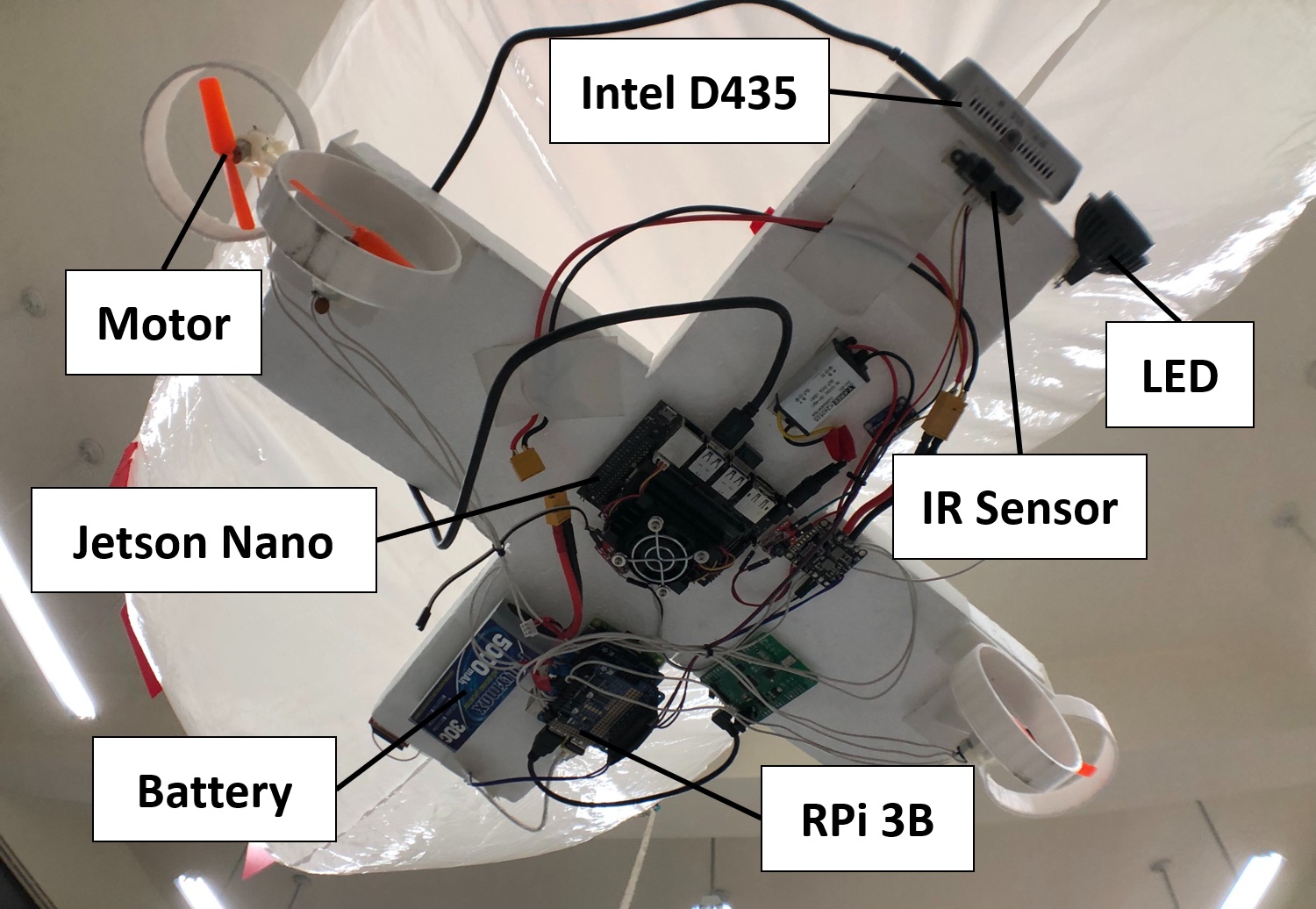}
\centering
\caption{Hardware Schematic Diagram of Duckiefloat.}
\label{fig:hardware_component}
\vspace{-10pt}
\end{figure}


Given the constrained passages in subterranean environments, the proposed blimp system is resource-constrained. A survey~\cite{berger2014comparison} is done about the hardware and software design perspective of resource constrained robots. One common trait is that these type of robots has a separation between low/high-level functions. For instance, central processing units are responsible for high-level tasks like communication, mapping and planning, and micro-controllers are connected with sensors and actuators to perform low-level tasks. For hardware design, power consumption, communication and robustness for maneuverability are also very critical since they are the basic functions of robots. Our designs and functionalities (lane/tunnel following and altitude controls) are inspired by the low-cost, resource constrained multi-robot platforms Duckietown~\cite{paull2017duckietown} and Duckiefly~\cite{brand2018pidrone}.




\section{System Descriptions}\label{system-descriptions}


Our goal is to develop an autonomous blimp systems, Duckiefloat, for search and rescue task in DARPA Subterranean Challenge. We summarize the requirements from the qualification and actual competition as follows:
 
\subsection{Requirements} \label{Requirements}

\subsubsection{Autonomous navigation in uncertain environments}

The robot can perform takeoff, landing, and is able to traverse through the course including at least two ninety-degree turns. The robot should also pass a constrained passage smaller than a certain dimensions.

\subsubsection{Perception in low-light or dusty environments}

The robot should autonomously identify artifacts with its own on-board lighting while navigating in no-light environments. It is also needed to carry out accurate localization (smaller than 5 meters deviations from ground truth).

\begin{table}[!t]
 \vspace{0.2cm}
 \footnotesize
 \caption{Duckiefloat vs. Quadcopter}
 \begin{tabular}{l l c c}
 \textbf{}          & \textbf{}         & \textbf{Duckiefloat}               & \textbf{Quadcopter}\cite{sa2017build}   \\

\toprule

 \textit{Dimension} & Body              & $120 \times 120 \times 250 $      & 65 (diagonal)         \\
 \textit{(cm)}      & Propeller         & 8 (diameter)  & 33 (diameter) \\
\midrule
 \textit{Payload}   & Floating          & 1600          & -         \\
 \textit{(g)}       & Max Takeoff       & 1800          & 3600      \\
\midrule
 \textit{Flight Time}   & No LED        & 60 - 90       & 16-40     \\
 \textit{(min)}         & With LED      & 40 - 50       & -         \\
\midrule
 \textit{Cost (USD)}      & Price        & 1,620         & 3,170     \\
\bottomrule

 \end{tabular}
 \label{table:HardwareCompare}

 \end{table}
\subsubsection{Long-term operations with limited computation/power resources}

The duration of each competition run is 1 hour for the tunnel circuit, and will be 1-2 hours in the urban and cave circuits. The power should provide flight time with sufficient payload, mobility, and on-board illumination. 

\subsubsection{Communications and supervisor interference}

Maintaining communication is ideal in order to submit artifact reports. While the communication bandwidth is available, mapping capability is useful for human supervisor to gain situational awareness. 

 \begin{table}[!b]
 \vspace{0.2cm}
 \footnotesize
 \caption{Duckiefloat Component Weights}
 \begin{tabular}{l P{4.5cm} r c}
 \textbf{Subsystem}        & \textbf{Item} & \textbf{Weight} \\

\toprule

 \textit{Sensors}          & Intel D435, Infrared Sensor         & 117g \\ 
 \textit{Computation}      & Raspberry Pi 3 B, Jetson Nano       & 248g \\
 \textit{Actuation}        & 4 DC motors, Shell                  & 172g \\
 \textit{Communication}    & LoRa module, XBee module            & 92g  \\
 \textit{Motor Controller} & Adafruit DC motor hat               & 50g  \\
 \textit{Power}            & 7.4V 5500 mAh Li-PO Battery         & 227g \\
 \textit{Body}             & Envelope, Platform, Tails           & 150g \\
 \textit{Illumination}     & 6W LED light                        & 50g  \\
 \textit{Waterproof}       & Waterproof tape                     & 120g \\
 \textit{Other}            & wires, adhesives, converters, tags  & 217g \\
\midrule
 \textit{Total}            &                                     & 1,423g \\
\bottomrule

 \end{tabular}
 \label{table:DuckiebotMinimal}
\footnotesize 
Weights are then added to make Duckiefloat slightly heavier than air.
\end{table}

Table~\ref{table:HardwareCompare} lists the comparisons between Duckiefloat and the commercially available quadcopter drone described in~\cite{sa2017build}, including payload, maximum takeoff weights, flight time, and etc. The flight time of Duckiefloat is designed to be longer than 60 minutes. Nevertheless, we found the onboard illumine and communication module consume significant battery life.


\subsection{Assumptions}

The vehicle's motion control is divided to altitude controls and planar movements. They are assumed as independent motions. The motor configuration is a basic differential drive, allowing linear and angular accelerations. We also assume the environmental airflow is limited. The ideal tunnel dimensions are more than 2 meters in height, and approximate 3 meters wide.

\subsection{System Overview}

We design Duckiefloat with dimensions of $1.2m \times 1.2m \times 2.5m$, which fits the assumed tunnels. Filled with helium gas, it carries around 1,600 grams, which is able to carry all electronic devices and batteries(Table.~\ref{table:DuckiebotMinimal}).

The system overview is shown in Fig.~\ref{fig:software}.
We used a Raspberry Pi 3B with an Adafruit DC Motor Hat as the motor controller. A NVIDIA Jetson Nano is the main computing unit which is responsible for perception and other high-level tasks. Furthermore, Duckiefloat is equipped with an infrared sensor that measures the altitude of the blimp.

We designed a cross-shaped Styrofoam platform, attached under the blimp to carry sensors, motors, and controllers, shown in Fig.~\ref{fig:hardware_component}. The baseline functionalities are tunnel following and artifact searching in dark environments. Therefore we need a camera which contains depth information that can also be able to work in low light conditions, and also combined the light weight and low computation requests. We used Intel RealSense D435 depth camera as our main sensor. At the same time, the Intel RealSense D435 also provides us data for visual odometry in order to build our map and localize the robot in the tunnel. 


\subsection{Mobility}

\subsubsection{Challenges}

Blimp has unique dynamics than other aerial vehicle, and has more
in common with submarine robots due to the buoyancy, mass effects, and their aerodynamics. 

\subsubsection{Algorithms: Altitude Controls}

We used PID controller to perform the altitude control, and set the goal altitude to 0.6 meter, which Duckiefloat would fly on a constant altitude.

\subsubsection{Algorithms: Tunnel Following}
In the tunnel circuit, we implement a baseline tunnel following policy for the blimp to perform exploration. By analyzing the pointcloud gathered from the RGB-D camera, we find the points of interest and project them to the plane parallel to the ground. Then we search line segments in the image and classify them into right, left or front wall. The slopes and intercepts of lines of different wall classes are then interpreted to a state of the blimp. The robot state at time $t$ is represented as $x_{t} = \langle d_{t}, \phi_{t} \rangle$,
where $d_{t}$ is the lateral distance between the blimp and the center of the tunnel at time $t$ and $\phi_{t}$ is the angle relative to the tunnel axis. Then a PID controller is used to control the robot state with target $d=0$  and $\phi=0$ which makes the blimp stay at the center of the tunnel with its yaw angle parallel to the tunnel.


\subsubsection{Implementations}

We designed the propulsion system with two DC motors to control the horizontal movements, and another two DC motors to control the vertical movements. The two sets of DC motors are perpendicular to each other.

\subsection{Perception}

\subsubsection{Challenges}

Lighting in these subterranean terrains are very limited. Only some light are emitted from the vehicle itself. So it is crucial for the algorithm have some robustness to different lighting conditions. The deep learning algorithm is also constrained by the computation and memory constraint of embedded boards.

\subsubsection{Algorithms}

In our specific task for DARPA SubT Challenge, we use SSD~\cite{liu2016ssd} to perform artifact searching. However computation and memory constraints limit the capability to run full size deep learning models. Therefore, Mobilenet-SSD~\cite{howard2017mobilenets} with 1GB of model size are perfect for our case.

\subsubsection{Implementations}

The main computation unit on Duckiefloat is Nvidia Jetson Nano, an embedded board with on-board GPU with 4GB shared memory and comparably low power consumption (10 watt) to other boards with a GPU. For the artifact search of the SubT challenge defined 5 classes of objects, including survivors, backpacks, cellphones, drills ,and fire-extinguishers. For training data we self-gathered 1300 images per class including 5 different view angles, 3 different view distances and different illumination conditions. 

\subsection{Communication}

\subsubsection{Challenges}

Communication is the key challenge in the competition. The multi-path problem and dimensions and depths of the tunnels affect the communication performances.  

\subsubsection{Implementations}

In order to have a longer range of communication in underground tunnel environments, we integrated the LoRa~\cite{vangelista2015long} communication module on Duckiefloat. The module provides long-range and low-power peer-to-peer communication between the base station and Duckiefloat. 
It allows modification on the bandwidth which affects data rate and communication range. We choose the bandwidth to be around 125KHz which the low data rate is an acceptable trade-off for a better communication range.
The low data rate is sufficient to update the robot state and a sparse 2D range image at the base station at 1Hz.

\subsection{Autonomy}

\subsubsection{Challenges}

Mapping and localization are harder to achieve in these underground environments due to the roughness of terrains and sometimes lack of distinct features (textures) which may cause failures for off-the-shelf SLAM algorithms.  


\subsubsection{Algorithms}

We perform ORB-SLAM~\cite{mur2017orb} algorithm for on-board visual odometry. During our on-site field tests, we found it vulnerable to fast movements (motion blur under low light) or illumination changes. Therefore we perform fail-safe mechanisms to detect fail cases(ie. visual odometry is lost) and recover to previous state. 

\subsubsection{Implementations}

In addition to the fail-safe mechanism, we consider that the system is not required to be fully autonomous in the challenge, as long as the communication maintains. We implement the minimal data representing the robot's surroundings. We take a slice of point cloud gathered by RGB-D camera and divide the angle of field of view into 8 bins. Then we only send the closest points in each of the 8 bins, resulting with 8 points(or less if no points in some bin). These points are sent to the base station at 1 Hz, providing the human supervisor some form of situational awareness. Then the base station could send moving signals to Duckiefloat if recovery is needed.

\begin{figure}[!hb]
\includegraphics[width=0.6\columnwidth]{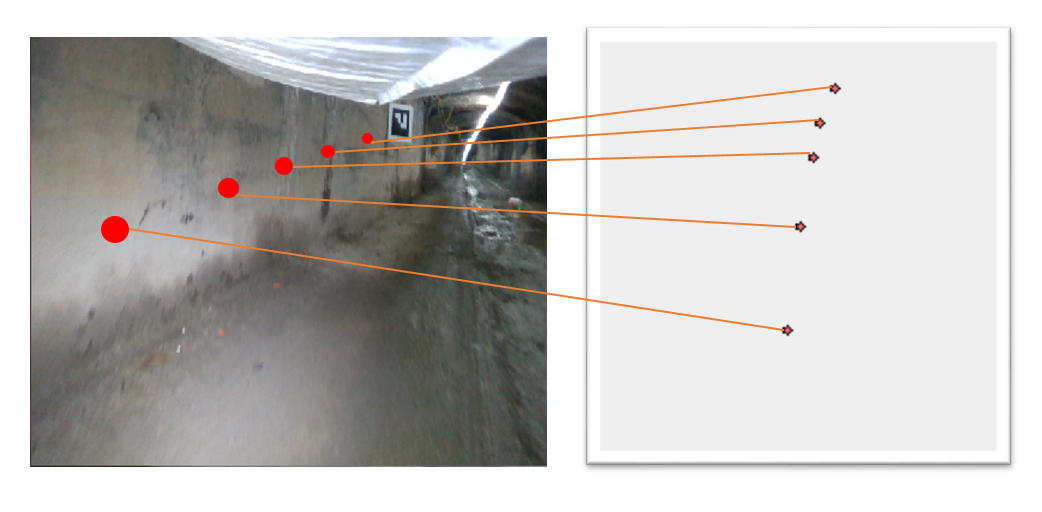}
\centering
\caption{Left: Duckiefloat camera view. Right: Points representing situational awareness}
\label{fig:tunnel-following}
\vspace{-10pt}
\end{figure}


\section{Experiments} \label{Experiments}
Several experiments are designed to test different aspects of Duckiefloat.

\begin{itemize}
	\item Autonomous vs. Human Controlled
	\item Situational Awareness and Failure Recovery
	\item Artifact Search
\end{itemize}


\subsection{Autonomous vs Human Controlled}
Most current robots designed for search and rescue operations have little or no autonomy at all, but instead they rely on operators remotely controlling them. However, having full control of the robot means having a very stable and reliable communication. 

\begin{figure}[!h]
\includegraphics[width=1.0\columnwidth]{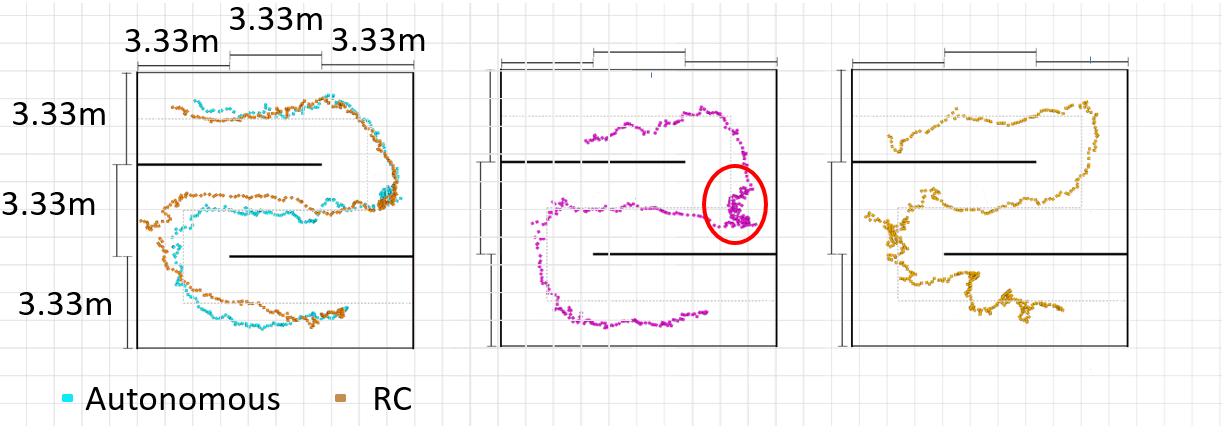}
\centering
\caption{Left: The S-shaped testing track setup. Trajectories of autonomous tunnel following and RC-controlled. Middle: Red circle shows human recovery Right: Turbulence causes collisions.}
\label{fig:Autonomous_vs_RC}
\end{figure}

We setup a testing field in the room with a s-shaped track, with one right and left u-turn. The path is around 3.3 meters wide. 
The room setup is as Fig.~\ref{fig:Autonomous_vs_RC}. (left)

In this experiment we want to show the differences between autonomous navigation and human tele-operated. The autonomous runs are based on the algorithms described in Sec.III part D. For human tele-operation, a stable 5GHz WiFi connection is established among Duckiefloat and the base station. The human supervisor stays at the base station and sends control signals to Duckiefloat according to the RGB image feed. Five runs are conducted for each remote-controlled (RC) and autonomous mode.
Ultra-wideband (UWB) localization system is installed in the room and on Duckiefloat to provide a reference trajectory in this experiment. Ultra-wideband (UWB) localization system has a mean absolute error of around 9.4 cm and an average standard deviation of 2 cm~\cite{Conceio2018RobotLI}. 5 points on the middle line of the track is picked for trajectory bench-marking. The lateral distance between the reference points and the trajectory is calculated then averaged.


The results are shown in Table.~\ref{table:rc-auto-results}. The results are calculated by averaging 5 runs. Both of the different methods have around 0.3 meters of trajectory error but the autonomous runs have a significant bigger standard deviation. This could be interpreted that a human operator may have more consistent driving path, especially turning u-turns. RC runs also finish the runs ~10 seconds faster than autonomous runs. It is quite obvious that human supervisors can take advantage from their experiences for better velocity control. Lastly, RC runs have less collision occurring than autonomous runs. Note: We only consider the autonomous runs that successfully travels through the whole track.

\begin{table}[!h]
\centering
\footnotesize
\caption{Experiment Results}
 \begin{tabular}{l c c c}

 \textbf{} & \textbf{RC} & \textbf{Auto} & \textbf{Auto w/ airflow} \\

\toprule

 \textit{Trajectory Error(m)}   & $0.31\pm0.15$ & $0.31\pm0.81$ & $0.07\pm0.05$   \\
 \textit{Duration(sec)}         & 72.6          & 85.05         & 112.0 \\
 \textit{Collision Count}       & 4.0           & 5.2           & 7.7   \\
\textit{Number of Runs}       & 5           & 5           & 3   \\

\bottomrule

 \end{tabular}
 \label{table:rc-auto-results}
\footnotesize 
\end{table}


From the overall performance, the RC method slightly outperforms the autonomous method. However, it is only possible to perform fully-RC if robots are in a controlled environment which has stable and reliable wireless connection. 

\subsection{Situational Awareness and Failure Recovery}
In real world scenarios, in particularly subterranean environments, communication is usually unreliable and unstable. However, we could still leverage the communication which has low bandwidth and high latency to provide the operator some information about the blimp robot and even send signals in order to recover the robot if its fails at some point.
From the previous experiments there are actually quite a few runs that Duckiefloat gets stuck at corners. In these cases we try to test whether the human supervisor can recover it only by reading situational awareness information at a low frequency(1 Hz). Among the 29 corners Duckiefloat encountered, 25(86\%) corners can be traversed by autonomous mode. 3(10\%) of the remaining 4 corners can be recovered by human intervention, 1(0.3\%) can't be recovered. Human recovering actions are mostly comparably complex
. Fig.~\ref{fig:Autonomous_vs_RC} shows some cases of human recovering, including strong turns and even backing up. For the cases human couldn't even recover, Duckiefloat is stuck to deep on some structures.

To prove the usability of the failure recovery system, we let Duckiefloat operate in the corridors of our building.

\begin{figure}[!h]
\includegraphics[width=0.8\columnwidth]{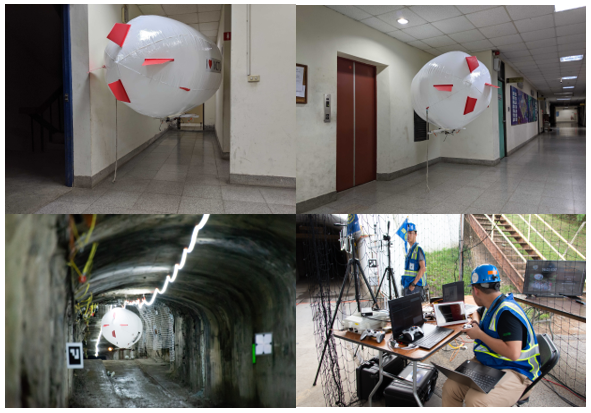}
\centering
\caption{Testing in real-world environments. Top: Scenarios Duckiefloat got stuck and could be recovered by human. Bottom left: Duckiefloat at NIOSH tunnel in the DARPA SubT Challenge. Bottom right: Human supervisor was monitoring situational awareness at base station.}
\label{fig:subt}
\vspace{-5pt}
\end{figure}

We let it run for as long as possible and with the help of human recovery constantly exploring new areas. As a result Duckiefloat operated for 47 minutes and covered around 500 meters, including going up the stairs autonomously. In the run, there are 8 situations that human recovery is needed. Those cases include corners and constrained paths(Fig.~\ref{fig:subt}).


\subsection{Artifacts Search}

In the aspect of artifacts search, we compared the performance of MobileNet-SSD \cite{liu2016ssd} and Tiny-YOLO \cite{redmon2016You} network. We trained both network with the exact same set of training data and chose the model with the best performance in training epochs. From our self-gathered test data, Tiny-YOLO and MobileNet-SSD resulted an AP(IOU=0.5) of 0.48 and 0.74 respectfully. We found out that Tiny-YOLO is much more sensitive to smaller objects or objects further from cameras which covered smaller pixel region in images. This gives the advantage of finding smaller objects like cellphones but on the other hand generates a lots more false positive results. We also found out that Tiny-YOLO from time to time classified objects into wrong classes.
\section{Real Environment Tests} \label{Real World Tests}
Previous experiments are all carried out in indoor and controlled environments in order to get better knowledge of the system. However, the robot system is designed to operate in real subterranean environments. We tested the system in two tunnel environments: Houli Tunnel and The National Institute for Occupational Safety and Health (NIOSH) coal mine tunnel.

Houli Tunnel is an old train tunnel built in 1908. The total length of the tunnel is 12.69 kilometers with approximately 4m in width. The tunnel is structure in a large long carve with no branch road along the way. Our Duckiefloat is tested in the tunnel and succeeded to operate for an average of one hour which travelled about 300 meters in each fully autonomous run.


The DARPA SubT Challenge is held at NIOSH, Pittsburgh. Two mine tunnels are maintained for research purposes. The tunnels extend many kilometers in length and include highly constrained passages with a rough and muddy floor. 40 artifacts are placed randomly in the tunnels. We deployed our blimp robot in the tunnels but unfortunately the constrained paths limited our blimp to perform the searching task. Duckiefloat is stuck at some places and our system successes to recover Duckiefloat and pilot it back to the mine entrance. 

\begin{figure}
\centering     
\subfigure[Houli Tunnel]{\label{fig:a}\includegraphics[width=0.33\columnwidth]{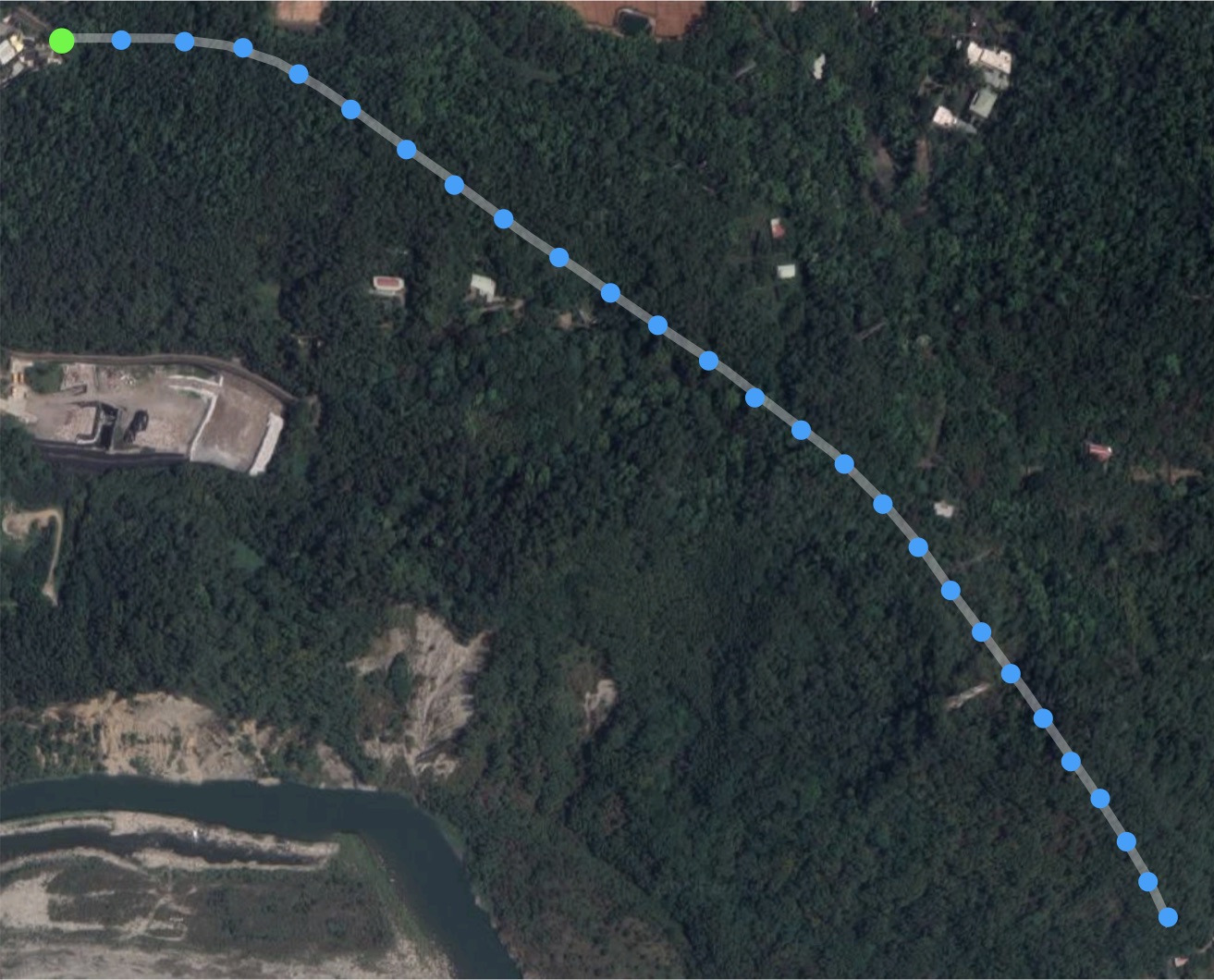}}
\subfigure[EE Building]{\label{fig:b}\includegraphics[width=0.26\columnwidth]{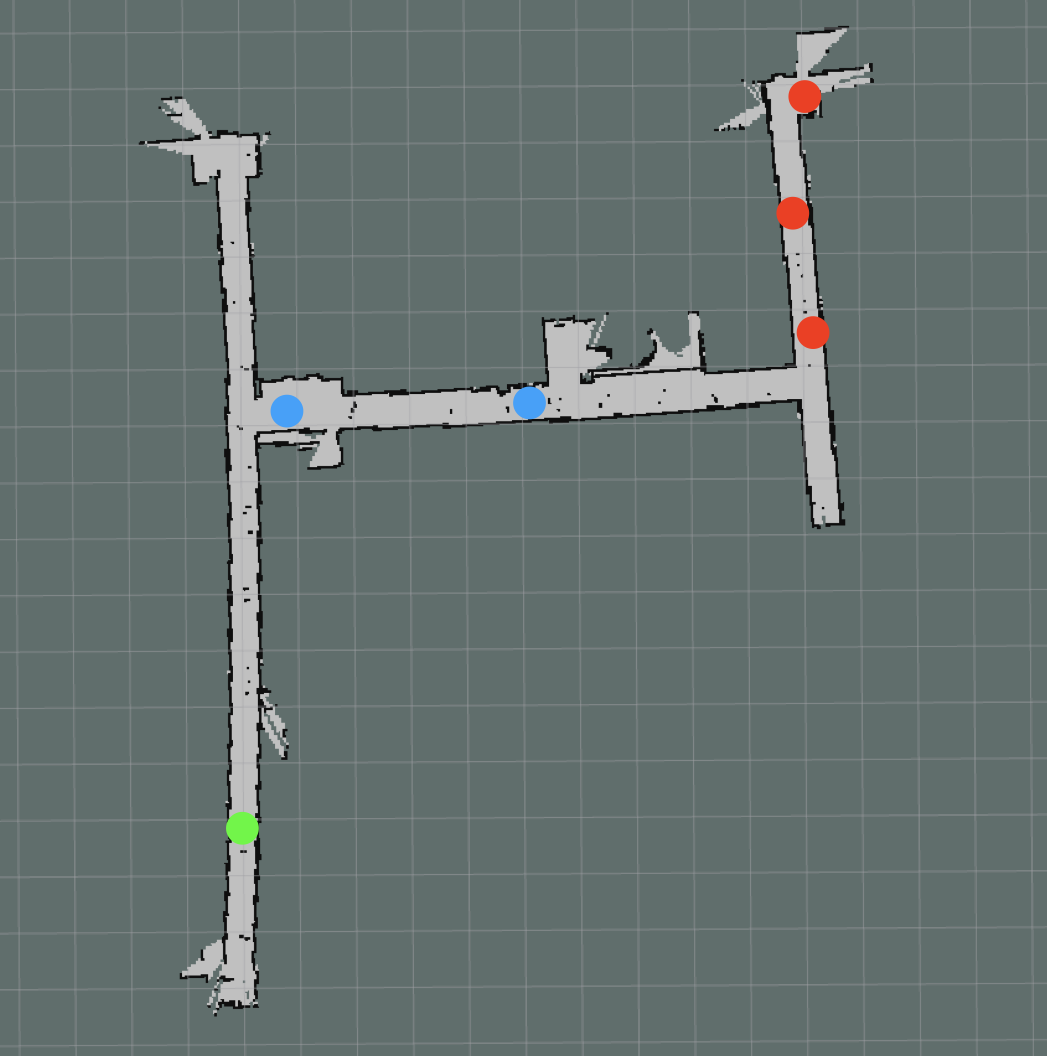}}
\subfigure[Air-raid Shelter]{\label{fig:b}\includegraphics[width=0.37\columnwidth]{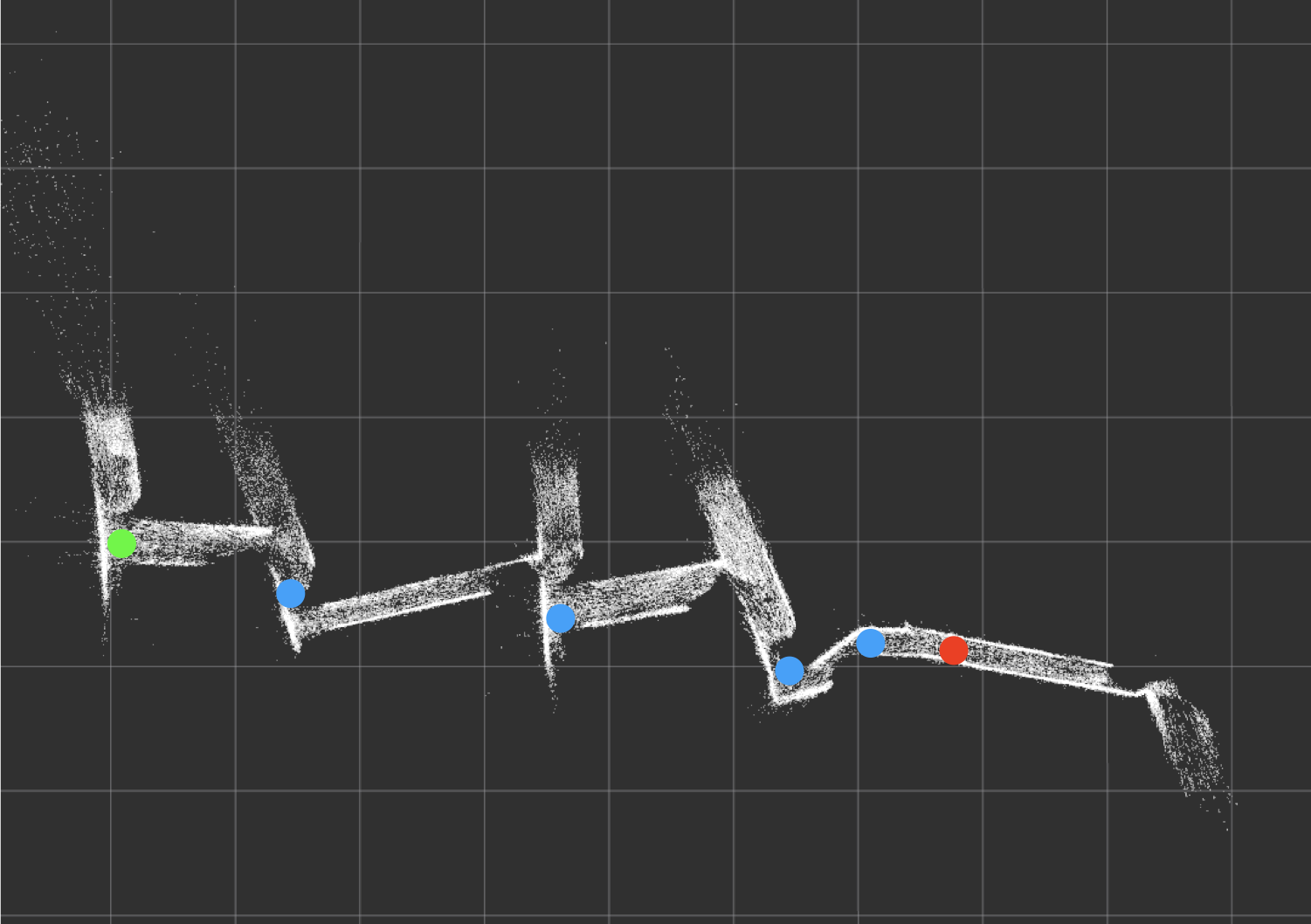}}

\subfigure[Comparison in Distance ]{\label{fig:lora_distance}\includegraphics[width=0.49\columnwidth]{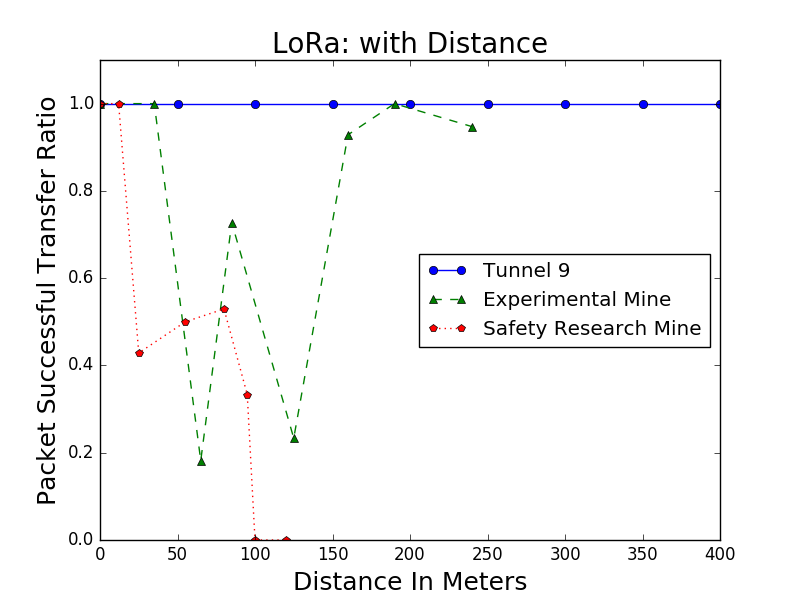}}
\subfigure[Comparison in Turns]{\label{fig:lora_turns}\includegraphics[width=0.49\columnwidth]{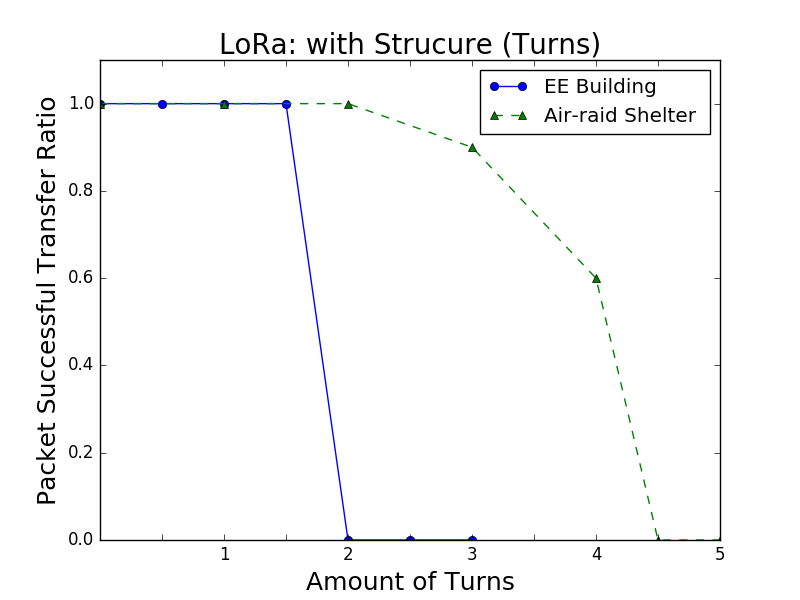}}
\caption{LoRa communication capability in five different tunnel or tunnel-like environments. To row shows our sampling points. Green dots are Base Stations and Red dots are locations that signals fail to reach Base Stations. (d) shows how distance affects LoRa signals in different environment. (e) shows how structure in different environments, i.e. turns, affects LoRa signals.}
\label{fig:lora}
\end{figure}

Communication is essential between Duckiefloat and Base Station in real world to provide situational awareness to human supervisors. We compared and assessed the communication capability of LoRa modules inside five different environments (Fig.\ref{fig:lora}). We analyzed the successful transfer ratio of LoRa packets within a time range as the capability of communication. We divided environments into two subsets. One set of them are those with longer paths which we analyzed how long that LoRa signal could travel within these environments. Other environments are smaller but more complicated which we analyzed how much turns LoRa signal could penetrate through. 

In the first set of environments, as Fig.\ref{fig:lora_distance} shows, LoRa signals can travel longer and with higher packets successful transfer ratio inside Houli Tunnel. We surmised that it is the size of Houli Tunnel makes the signal transfer easier. While three tunnels are roughly in same width, Houli Tunnel are 3 times higher than the other two tunnels. Another reason could be the material inside tunnels or within the wall of tunnels. Metal causes Shielding Effect which weaken penetrating power of electromagnetic field. Thus the metal structures inside tunnel crippled LoRa signals. 

In EE Building and the air-raid shelter, we analyzed amount of turns that signals can penetrate through (Fig.\ref{fig:lora_turns}). We defined a turn as the path turning right and left. Although EE Building has wider path and higher ceiling, the signal went through more turns inside the air-raid shelter. We surmised that the long lateral distance (50m) of the first turn of EE Building weakens signals. On the other hand the accumulative lateral distance of air-raid shelter is only 5 meter. Thus we concluded that if a turn happened, the larger the lateral distance the weaker LoRa signals would become while traversing through.






\section{Conclusions} \label{Conclusions}

Inspired by DARPA SubT Challenge, we developed a search and rescue blimp robot "Duckiefloat" with the abilities of performing SLAM and artifacts search via deep learning approach. Duckiefloat provide situational awareness so that human supervisors can gain information of unknown environments for intervention and failure recovery if necessary. We carried out experiments in real-world tunnel or tunnel-like environments with quantitative and qualitative results, providing insights for future directions of development search-purpose blimp robots.

 We performed experiments to compare the flying performance in control airflow (1.2m/s) and windless environments. The results indicate the number of collisions increase, and overall speed decreases. During our real world test in Houli Tunnel and NIOSH Tunnel, our Duckiefloat system encounter some turbulence, Duckiefloat got stuck by the airflow and hard to control. We will change the configurations of the motors or replace our current motors in the future.


\section*{Acknowledgments}
\label{sec:Acknowledgments}

The research was supported by Ministry of Science and Technology, Taiwan (grant 107-2923-E009-004-MY3, 108-2218-E-007-039, 108-2321-B-009-004).

\bibliographystyle{IEEEtran}
\bibliography{DuckietownBib.bib}

\begin{thebibliography}{10}
\providecommand{\url}[1]{#1}
\csname url@samestyle\endcsname
\providecommand{\newblock}{\relax}
\providecommand{\bibinfo}[2]{#2}
\providecommand{\BIBentrySTDinterwordspacing}{\spaceskip=0pt\relax}
\providecommand{\BIBentryALTinterwordstretchfactor}{4}
\providecommand{\BIBentryALTinterwordspacing}{\spaceskip=\fontdimen2\font plus
\BIBentryALTinterwordstretchfactor\fontdimen3\font minus
  \fontdimen4\font\relax}
\providecommand{\BIBforeignlanguage}[2]{{%
\expandafter\ifx\csname l@#1\endcsname\relax
\typeout{** WARNING: IEEEtran.bst: No hyphenation pattern has been}%
\typeout{** loaded for the language `#1'. Using the pattern for}%
\typeout{** the default language instead.}%
\else
\language=\csname l@#1\endcsname
\fi
#2}}
\providecommand{\BIBdecl}{\relax}
\BIBdecl

\bibitem{sa2017build}
I.~Sa, M.~Kamel, M.~Burri, M.~Bloesch, R.~Khanna, M.~Popovi{\'c}, J.~Nieto, and
  R.~Siegwart, ``Build your own visual-inertial drone: A cost-effective and
  open-source autonomous drone,'' \emph{IEEE Robotics \& Automation Magazine},
  vol.~25, no.~1, pp. 89--103, 2017.

\bibitem{SubT-website}
\BIBentryALTinterwordspacing
DARPA. (2019) Darpa subterranean challenge. [Online]. Available:
  \url{https://www.subtchallenge.com/}
\BIBentrySTDinterwordspacing

\bibitem{delmerico2019current}
J.~Delmerico, S.~Mintchev, A.~Giusti, B.~Gromov, K.~Melo, T.~Horvat, C.~Cadena,
  M.~Hutter, A.~Ijspeert, D.~Floreano \emph{et~al.}, ``The current state and
  future outlook of rescue robotics,'' \emph{Journal of Field Robotics}, 2019.

\bibitem{wang2014development}
W.~Wang, W.~Dong, Y.~Su, D.~Wu, and Z.~Du, ``Development of search-and-rescue
  robots for underground coal mine applications,'' \emph{Journal of Field
  Robotics}, vol.~31, no.~3, pp. 386--407, 2014.

\bibitem{gu2018robot}
Y.~Gu, J.~Strader, N.~Ohi, S.~Harper, K.~Lassak, C.~Yang, L.~Kogan, B.~Hu,
  M.~Gramlich, R.~Kavi \emph{et~al.}, ``Robot foraging: Autonomous sample
  return in a large outdoor environment,'' \emph{IEEE Robotics \& Automation
  Magazine}, no.~99, pp. 1--1, 2018.

\bibitem{niroui2019deep}
F.~Niroui, K.~Zhang, Z.~Kashino, and G.~Nejat, ``Deep reinforcement learning
  robot for search and rescue applications: Exploration in unknown cluttered
  environments,'' \emph{IEEE Robotics and Automation Letters}, vol.~4, no.~2,
  pp. 610--617, 2019.

\bibitem{montero2015past}
R.~Montero, J.~Victores, S.~Martinez, A.~Jard{\'o}n, and C.~Balaguer, ``Past,
  present and future of robotic tunnel inspection,'' \emph{Automation in
  Construction}, vol.~59, pp. 99--112, 2015.

\bibitem{chmelina2007development}
K.~Chmelina and A.~Maierhofer, ``Development of an underground construction
  information system in the eu research project tunconstruct,''
  \emph{Computational Modelling in Tunnelling}, 2007.

\bibitem{loupos2014robotic}
K.~Loupos, A.~Amditis, C.~Stentoumis, P.~Chrobocinski, J.~Victores, M.~Wietek,
  P.~Panetsos, A.~Roncaglia, S.~Camarinopoulos, V.~Kalidromitis \emph{et~al.},
  ``Robotic intelligent vision and control for tunnel inspection and
  evaluation-the robinspect ec project,'' in \emph{2014 IEEE International
  Symposium on Robotic and Sensors Environments (ROSE) Proceedings}.\hskip 1em
  plus 0.5em minus 0.4em\relax IEEE, 2014, pp. 72--77.

\bibitem{roberts2000autonomous}
J.~M. Roberts, E.~S. Duff, P.~I. Corke, P.~Sikka, G.~J. Winstanley, and
  J.~Cunningham, ``Autonomous control of underground mining vehicles using
  reactive navigation,'' in \emph{Proceedings 2000 ICRA. Millennium Conference.
  IEEE International Conference on Robotics and Automation. Symposia
  Proceedings (Cat. No. 00CH37065)}, vol.~4.\hskip 1em plus 0.5em minus
  0.4em\relax IEEE, 2000, pp. 3790--3795.

\bibitem{shaffer1992robotic}
G.~Shaffer and A.~Stentz, ``A robotic system for underground coal mining,'' in
  \emph{Proceedings 1992 IEEE International Conference on Robotics and
  Automation}.\hskip 1em plus 0.5em minus 0.4em\relax IEEE, 1992, pp. 633--638.

\bibitem{mur2017orb}
R.~Mur-Artal and J.~D. Tard{\'o}s, ``Orb-slam2: An open-source slam system for
  monocular, stereo, and rgb-d cameras,'' \emph{IEEE Transactions on Robotics},
  vol.~33, no.~5, pp. 1255--1262, 2017.

\bibitem{jacobson2018semi}
A.~Jacobson, F.~Zeng, D.~Smith, N.~Boswell, T.~Peynot, and M.~Milford,
  ``Semi-supervised slam: Leveraging low-cost sensors on underground autonomous
  vehicles for position tracking,'' in \emph{2018 IEEE/RSJ International
  Conference on Intelligent Robots and Systems (IROS)}.\hskip 1em plus 0.5em
  minus 0.4em\relax IEEE, 2018, pp. 3970--3977.

\bibitem{zeng2019lookup}
F.~Zeng, A.~Jacobson, D.~Smith, N.~Boswell, T.~Peynot, and M.~Milford,
  ``Lookup: Vision-only real-time precise underground localisation for
  autonomous mining vehicles,'' \emph{arXiv preprint arXiv:1903.08313}, 2019.

\bibitem{huber2003automatic}
D.~F. Huber and N.~Vandapel, ``Automatic 3d underground mine mapping,'' in
  \emph{Field and Service Robotics}.\hskip 1em plus 0.5em minus 0.4em\relax
  Springer, 2003, pp. 497--506.

\bibitem{xiong2009integrated}
C.~Xiong, D.~Han, and Y.~Xiong, ``An integrated localization system for robots
  in underground environments,'' \emph{Industrial Robot: An International
  Journal}, vol.~36, no.~3, pp. 221--229, 2009.

\bibitem{gonzalez2009developing}
P.~Gonz{\'a}lez, W.~Burgard, R.~Sanz~Dom{\'\i}nguez, and
  J.~L{\'o}pez~Fern{\'a}ndez, ``Developing a low-cost autonomous indoor
  blimp,'' 2009.

\bibitem{fedorenko2016indoor}
R.~Fedorenko and V.~Krukhmalev, ``Indoor autonomous airship control and
  navigation system,'' in \emph{MATEC Web of Conferences}, vol.~42.\hskip 1em
  plus 0.5em minus 0.4em\relax EDP Sciences, 2016, p. 01006.

\bibitem{rottmann2007towards}
A.~Rottmann, T.~Zitterell, W.~Burgard, L.~Reindl, C.~Scholl \emph{et~al.},
  ``Towards an experimental autonomous blimp platform,'' 2007.

\bibitem{ko2007gaussian}
J.~Ko, D.~J. Klein, D.~Fox, and D.~Haehnel, ``Gaussian processes and
  reinforcement learning for identification and control of an autonomous
  blimp,'' in \emph{Proceedings 2007 ieee international conference on robotics
  and automation}.\hskip 1em plus 0.5em minus 0.4em\relax IEEE, 2007, pp.
  742--747.

\bibitem{hygounenc2004autonomous}
E.~Hygounenc, I.-K. Jung, P.~Soueres, and S.~Lacroix, ``The autonomous blimp
  project of laas-cnrs: Achievements in flight control and terrain mapping,''
  \emph{The International Journal of Robotics Research}, vol.~23, no. 4-5, pp.
  473--511, 2004.

\bibitem{fukao2003autonomous}
T.~Fukao, K.~Fujitani, and T.~Kanade, ``An autonomous blimp for a surveillance
  system,'' in \emph{Proceedings 2003 IEEE/RSJ International Conference on
  Intelligent Robots and Systems (IROS 2003)(Cat. No. 03CH37453)},
  vol.~2.\hskip 1em plus 0.5em minus 0.4em\relax IEEE, 2003, pp. 1820--1825.

\bibitem{berger2014comparison}
C.~Berger and M.~Dukaczewski, ``Comparison of architectural design decisions
  for resource-constrained self-driving cars-a multiple case-study,''
  \emph{Informatik 2014}, 2014.

\bibitem{paull2017duckietown}
L.~Paull, J.~Tani, H.~Ahn, J.~Alonso-Mora, L.~Carlone, M.~Cap, Y.~F. Chen,
  C.~Choi, J.~Dusek, Y.~Fang \emph{et~al.}, ``Duckietown: an open, inexpensive
  and flexible platform for autonomy education and research,'' in \emph{2017
  IEEE International Conference on Robotics and Automation (ICRA)}.\hskip 1em
  plus 0.5em minus 0.4em\relax IEEE, 2017, pp. 1497--1504.

\bibitem{brand2018pidrone}
I.~Brand, J.~Roy, A.~Ray, J.~Oberlin, and S.~Oberlix, ``Pidrone: An autonomous
  educational drone using raspberry pi and python,'' in \emph{2018 IEEE/RSJ
  International Conference on Intelligent Robots and Systems (IROS)}.\hskip 1em
  plus 0.5em minus 0.4em\relax IEEE, 2018, pp. 1--7.

\bibitem{liu2016ssd}
W.~Liu, D.~Anguelov, D.~Erhan, C.~Szegedy, S.~Reed, C.-Y. Fu, and A.~C. Berg,
  ``Ssd: Single shot multibox detector,'' in \emph{European conference on
  computer vision}.\hskip 1em plus 0.5em minus 0.4em\relax Springer, 2016, pp.
  21--37.

\bibitem{howard2017mobilenets}
A.~G. Howard, M.~Zhu, B.~Chen, D.~Kalenichenko, W.~Wang, T.~Weyand,
  M.~Andreetto, and H.~Adam, ``Mobilenets: Efficient convolutional neural
  networks for mobile vision applications,'' \emph{arXiv preprint
  arXiv:1704.04861}, 2017.

\bibitem{vangelista2015long}
L.~Vangelista, A.~Zanella, and M.~Zorzi, ``Long-range iot technologies: The
  dawn of lora™,'' in \emph{Future Access Enablers of Ubiquitous and
  Intelligent Infrastructures}.\hskip 1em plus 0.5em minus 0.4em\relax
  Springer, 2015, pp. 51--58.

\bibitem{Conceio2018RobotLI}
T.~Conceiç{\~a}o, ``Robot localization in an agricultural environment,'' 2018.

\bibitem{redmon2016You}
J.~{Redmon}, S.~{Divvala}, R.~{Girshick}, and A.~{Farhadi}, ``You only look
  once: Unified, real-time object detection,'' in \emph{2016 IEEE Conference on
  Computer Vision and Pattern Recognition (CVPR)}, June 2016, pp. 779--788.

\end{thebibliography}

\end{document}